\documentclass[journal]{IEEEtran}
\usepackage{subfigure}
\usepackage{amsmath,amsfonts}
\usepackage[linesnumbered,ruled,vlined]{algorithm2e}
\usepackage{array}
\usepackage[caption=false,font=normalsize,labelfont=sf,textfont=sf]{subfig}
\usepackage{textcomp}
\usepackage{stfloats}
\usepackage{url}
\usepackage{verbatim}
\usepackage{graphicx}
\usepackage{cite}
\usepackage{subfig}
\usepackage{indentfirst}
\usepackage{graphicx}   

\usepackage{titlesec}

\begin{document}
\title{DNN Task Assignment in UAV Networks: A Generative AI Enhanced Multi-Agent Reinforcement Learning Approach}


\author{Xin Tang, Qian Chen, Wenjie Weng, Binhan Liao, Jiacheng Wang, Xianbin Cao, Xiaohuan Li

\thanks{This work was supported in part by the National Natural Science Foundation of China under Grant U22A2054, in part by the Guangxi Natural Science Foundation of China under Grant 2024JJA170165, and in part by the Graduate Study Abroad Program of Guilin University of Electronic Technology under Grant GDYX2024001. ({\itshape Corresponding author: Xiaohuan Li}.)}
\thanks{Xin Tang is with the Guangxi University Key Laboratory of Intelligent Networking and Scenario System (School of Information and Communication, Guilin University of Electronic Technology), Guilin 541004, China, and also with the College of Computing and Data Science, Nanyang Technological University, 639798, Singapore (e-mails: tangx@mails.guet.edu.cn).}
\thanks{Qian Chen is with the School of Architecture and Transportation Engineering, Guilin University of Electronic Technology, Guilin 541004, China (e-mails: chenqian@mails.guet.edu.cn).}
\thanks{Wenjie Weng, Binhan Liao and Xiaohuan Li are with the Guangxi University Key Laboratory of Intelligent Networking and Scenario System (School of Information and Communication, Guilin University of Electronic Technology), Guilin 541004, China, and also with National Engineering Laboratory for Comprehensive Transportation Big Data Application Technology (Guangxi), Nanning 530001, China (e-mails: wwjdzsyx@163.com; binhanliaoguet@163.com; lxhguet@guet.edu.cn).}
\thanks{Jiacheng Wang is with the College of Computing and Data Science, Nanyang Technological University, 639798, Singapore (e-mail: jiacheng.wang@ntu.edu.sg).}
\thanks{Xianbin Cao is with the School of Electronic and Information Engineering and the Key Laboratory of Advanced Technology of Near Space Information System, Ministry of Industry and Information Technology of China, Beihang University, Beijing 100191, China (e-mail: xbcao@buaa.edu.cn).}
}

\maketitle

\begin{abstract}
Unmanned Aerial Vehicles (UAVs) offer high mobility and flexible deployment capabilities, making them ideal for Internet of Things (IoT) applications. However, the substantial amount of data generated by various applications within the existing low-altitude economic network requires processing through Deep Neural Networks (DNN) on UAVs, which is challenging due to their limited computational resources. To address this issue, we propose a two-stage optimization method for flight path planning and task allocation based on a mother-child UAV swarm system. In the first stage, we employ a greedy algorithm to solve the path planning problem by considering the task size of the target area to be inspected and the shortest flight path as constraints. The goal is to minimize both the flight path of the UAV and the overall cost of the system. In the second stage, we introduce a novel DNN task assignment algorithm that combines Multi-Agent Deep Deterministic Policy Gradient (MADDPG) and Generative Diffusion Models (GDM), named GDM-MADDPG. This algorithm takes advantage of the reverse denoising process of GDM to replace the actor network in MADDPG. It enables UAVs to generate specific DNN task assignment actions based on agents' observations in a dynamic environment, thereby improving the efficiency of task scheduling and overall system performance. The simulation results demonstrate that our algorithm outperforms the benchmarks in terms of path planning, Age of Information (AoI), task completion, and system utility, demonstrating its effectiveness.
\end{abstract}


\section{Introduction}
\begin{figure}[!t]
	\centering
	\includegraphics[width=3.5in]{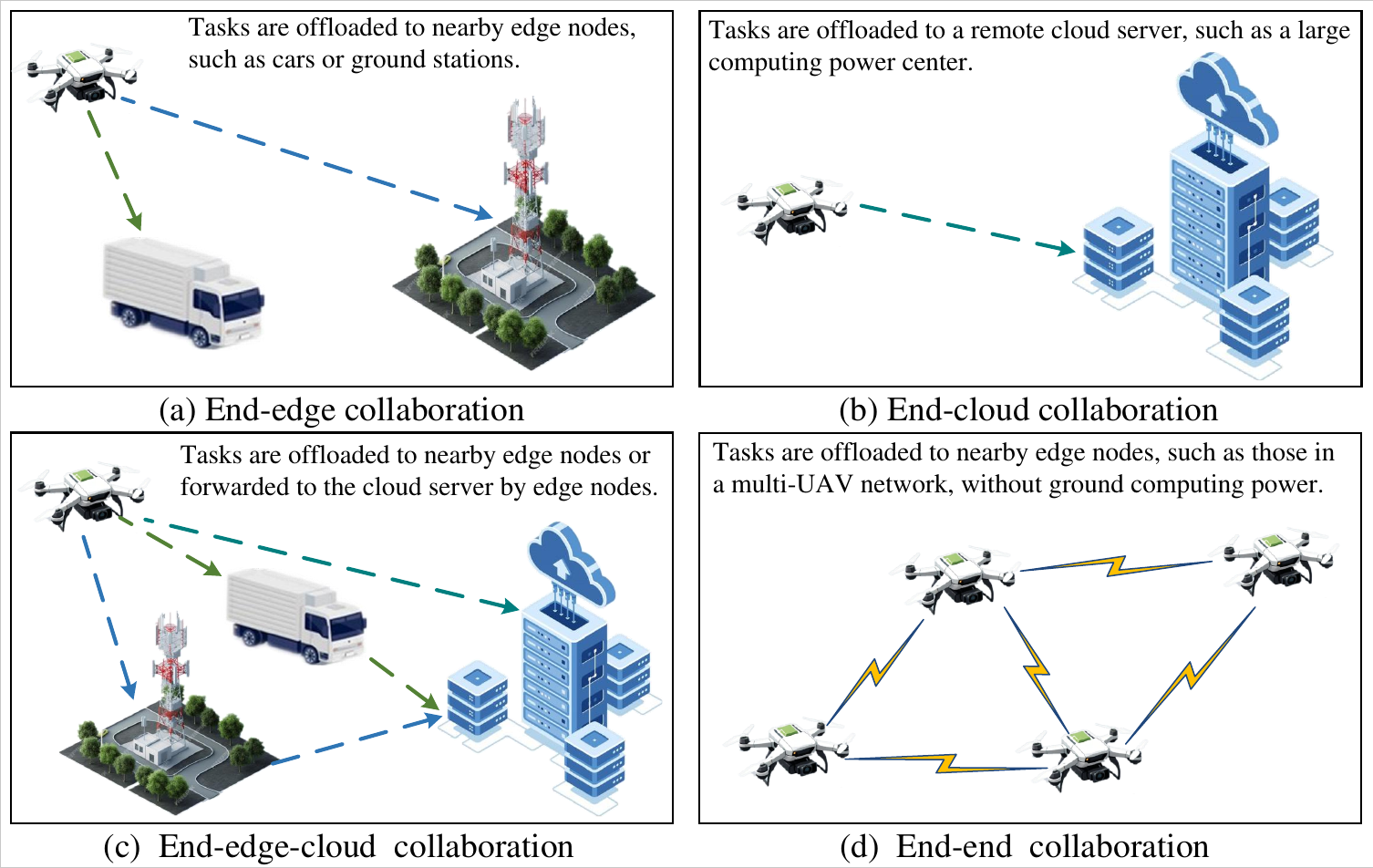}%
	\caption{Four task assignment models.}
	\label{fig01}
\end{figure}
Advancements in Unmanned Aerial Vehicles (UAVs) technology have led to their increasing use in various civil and military domains, including aerial object detection, rapid rescue operations in disaster-stricken areas \cite{raja2024ugen, wangT2023uav}, emergency scenarios \cite{tian2023uav}, and extensive agriculture \cite{xiong2024detecting}. Unlike traditional Internet of Things (IoT) devices \cite{gao2019dynamic}, UAVs offer not only lower costs, but also superior mobility and enhanced data collection capabilities \cite{wu2023split}. Although UAVs are well suited for the critical and complex tasks mentioned, managing the substantial data they generate remains a challenge \cite{ wang2021energy}. With the rise of artificial intelligence (AI), deep neural networks (DNN) have been proposed to handle the extensive data produced \cite{liu2024dnn}. Despite their efficiency in processing data, DNNs require significant resources due to their intricate structure and high performance demands. General methodologies segmented DNN models by layers, treating each layer as the primary computational unit, which often led to inefficiencies in collaborative inference due to the substantial computational demands of certain layers. 

Given the size and cost constraints of UAVs, their power supply is often limited and computing capacity is relatively low. Generally, there are two primary approaches to tackle this issue. One approach involves deploying lightweight models on UAVs, which reduces computational demands and inference latency but sacrifices accuracy \cite{sun2024all}. However, executing computation-intensive tasks independently with a single UAV presents significant challenges in achieving low latency and high inference accuracy. Another approach is to design from the perspective of the end-edge-cloud network architecture, enabling real-time, high-accuracy processing on UAVs with support from cloud and edge servers.

\begin{table*}[!t]
  \centering
  \caption{Comparison between related works and our proposed \label{tab:table01}} 
  \begin{tabular*}{\textwidth}{@{\extracolsep{\fill}}|c|c|c|c|c|c|c|c|} 
    \hline
    Ref. &Architecture &Adaptive partitioning& Path planning &Latency &Energy consumption &Stability &Age of information \\
    \hline
   \cite{deng2024integrated} &{End to edge}&{\checkmark}&{ }&{\checkmark}&{ }&  & \\
    \hline
    \cite{qu2023elastic} &{End to end}&{\checkmark}&{ }&{\checkmark}&{ }& {\checkmark} & \\
    \hline
    \cite{jouhari2021distributed} &{End to end}&{ }& {\checkmark} & {\checkmark} & &{\checkmark}&  \\
    \hline
    \cite{wang2023object} &{End to end / edge}& &{ }&{\checkmark}&{\checkmark}&{ }&  \\
    \hline
    \cite{zhao2022reliable} &{End to end}& &{ }& {\checkmark} & {\checkmark} & {\checkmark} &{ }\\
    \hline
    \cite{yang2023multi} &{End to end / cloud}&{\checkmark}&  &{\checkmark}&  & &  \\
    \hline
    \cite{long2024AoI} &{End to edge}&{ }& {\checkmark}& {} & {} & {} & {\checkmark}\\
    \hline
    \cite{ren2024efficient} &{End to end}& {}&{ } & {\checkmark} & {} & {} &{ }\\
    \hline
    \cite{lins2021artificial} &{End to end / edge}& { }&{\checkmark} &{\checkmark} &{ } &{ } & \\
    \hline
    \cite{sun2024all} &{End to end}& {\checkmark}&{ } &{\checkmark}  &{} &{ } &{ }\\
    \hline
    Proposed & {End to end} & {\checkmark} & {\checkmark} & {\checkmark} & {\checkmark} & {\checkmark} & {\checkmark} \\
    \hline
  \end{tabular*}
\end{table*}

Recent studies have shifted focus from relying solely on the cloud or a single edge device, such as a UAV, to harnessing the combined capabilities of both robust external cloud resources and edge servers for the collaborative execution of complex task. This collaborative approach can generally be categorized into four types of collaborative intelligence: End-edge, end-cloud, end-edge-cloud, and end-end \cite{qu2023elastic}. Fig. \ref{fig01} shows the four conventional paradigms. Although the first three types effectively mitigate extensive original data transmission to minimize latency and uphold high accuracy, such as DNN inference ability, they may still encounter significant processing latencys and potential failures when implemented in UAV contexts due to the instability of air-to-ground communication links, particularly in real-time dynamic applications. As end resources become increasingly abundant, some researchers have explored the possibility of conducting collaborative DNN inference across multiple end devices, specifically UAVs in the scenario under consideration, without dependence on edge servers or the cloud, termed end-end. This model aligns seamlessly with UAV operations, which typically involve task execution within UAV swarms, thereby reducing execution time and improving fault tolerance.

\subsection{Motivation and Challenges}

The primary challenge is that conventional scene data collection and monitoring tasks rely on staged task processing. For example, in \cite{deng2024integrated, wang2023uav}, the methods involve an initial phase of task planning, followed by data collection, and conclude with online or offline processing of related tasks, such as target detection and positioning. This sequential approach is evidently inadequate to meet the real-time requirements of sudden events or emergency scenarios. In addition, the limited computational power, battery life, and other onboard resources of UAVs further constrain their ability to perform real-time tasks effectively.

Secondly, when DNN tasks need to be offloaded, many studies explore the transfer of intermediate data to the ground nodes \cite{wang2023object, yang2023multi, lins2021artificial}. However, non-line-of-sight communication and the high relative mobility between UAVs and ground nodes—especially dynamic nodes—introduce significant uncertainty into the offloading process. Furthermore, existing methods for offloading computationally intensive tasks, such as heuristic-based, decomposition-based \cite{deng2024integrated} and game-based approaches \cite{gao2018game}, typically require numerous iterations to converge to a satisfactory and stable assignment decision. This iterative decision-making process is often impractical for latency-sensitive tasks and energy-constrained systems.

 \begin{figure*}[t!]
 	\centering
 	  \subfigure[Yolov5.]{
 \includegraphics[width=0.31\textwidth]{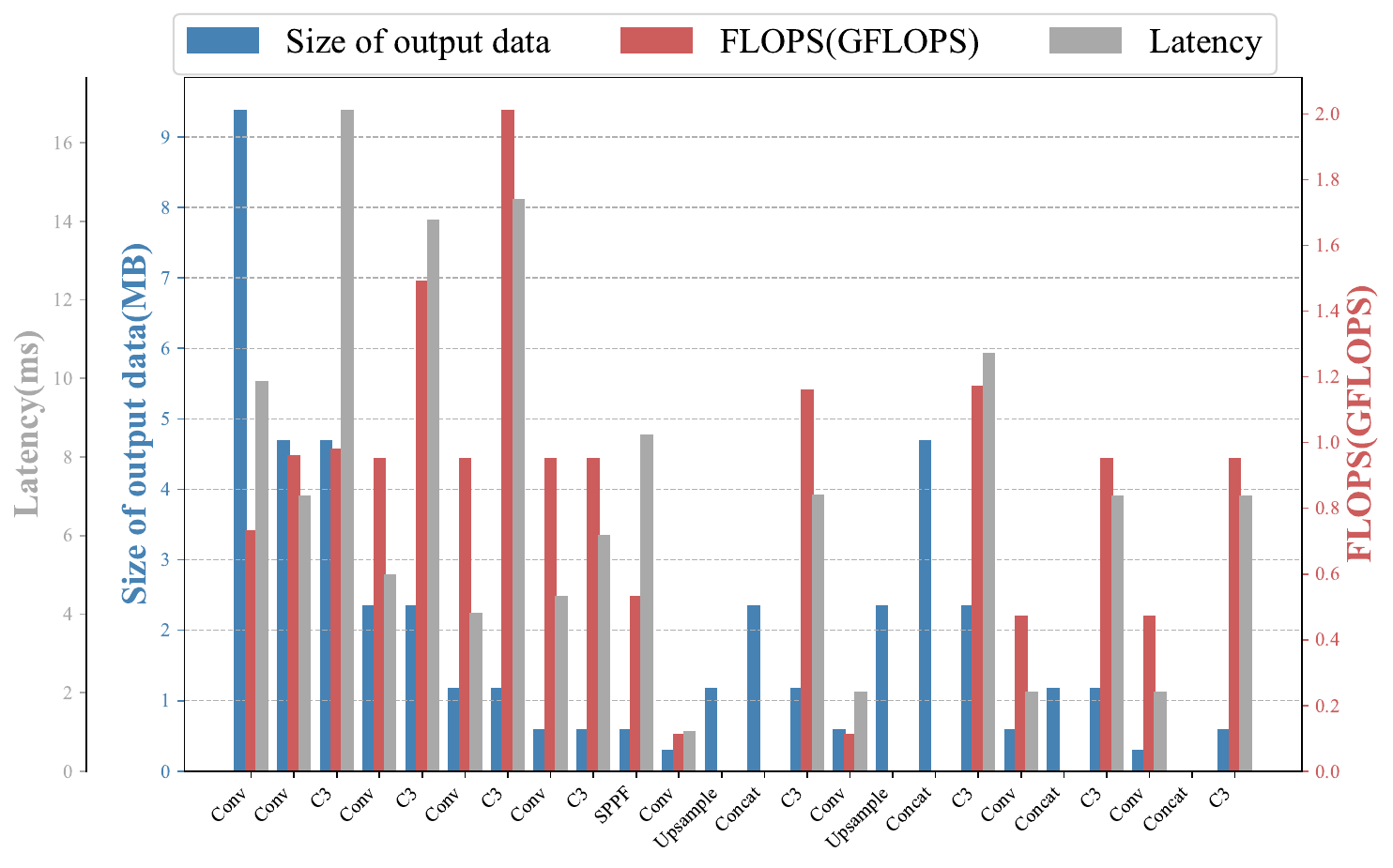}
 	}
        \hfill
         \subfigure[AlexNet.]{
 \includegraphics[width=0.31\textwidth]{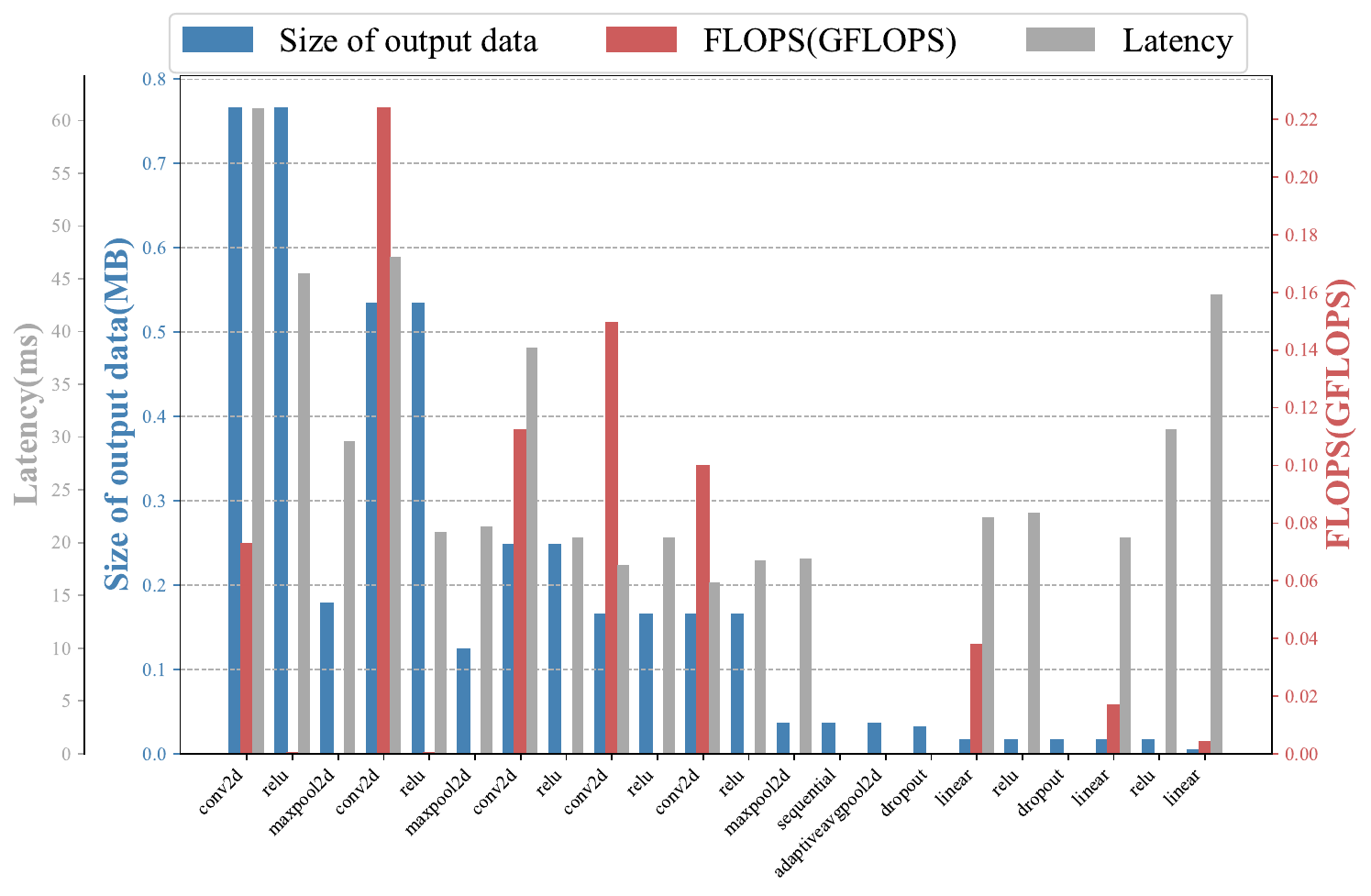}
 	}
        \hfill    
         \subfigure[VGG16.]{
 \includegraphics[width=0.31\textwidth]{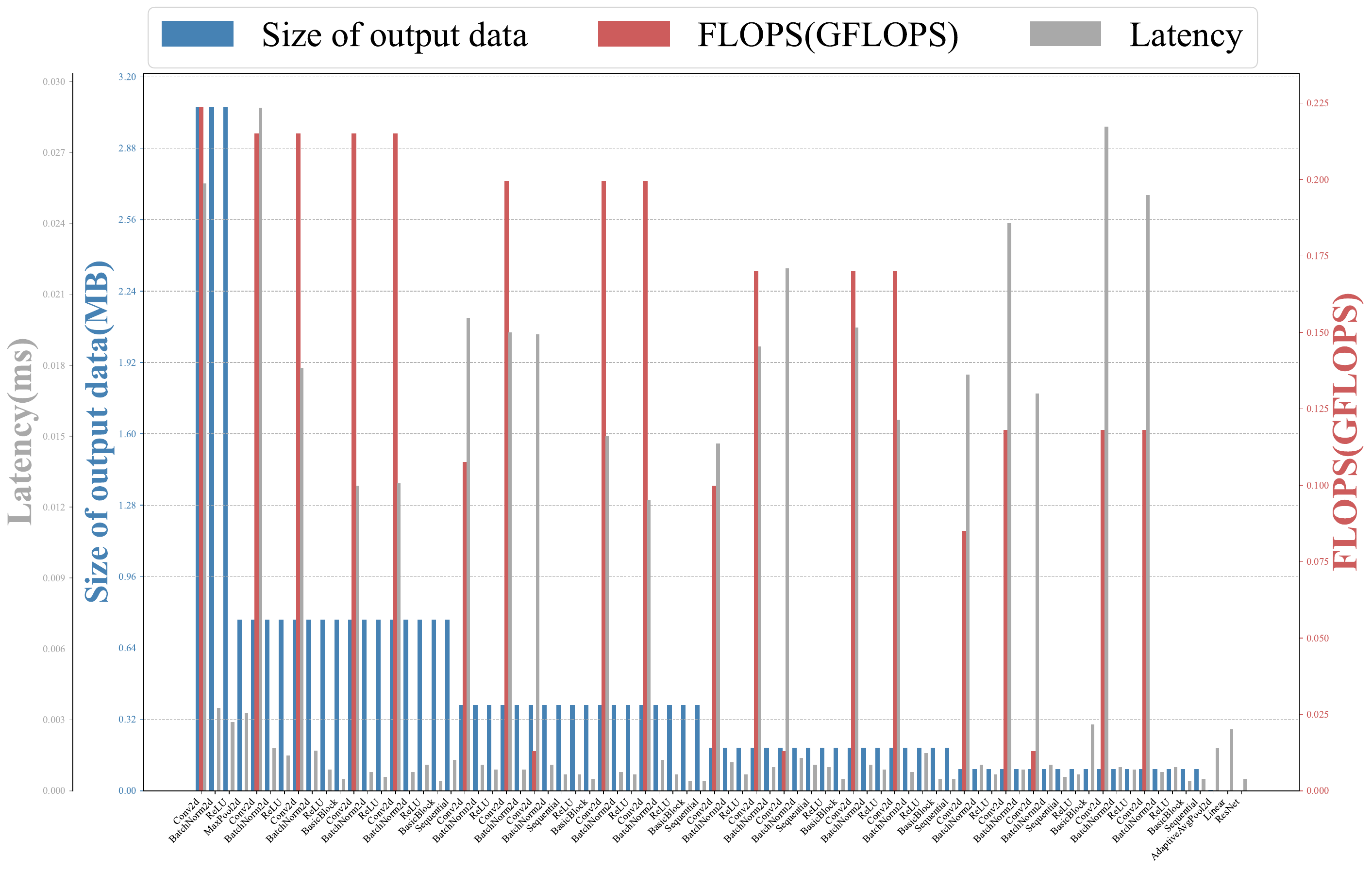}
 	}
 	\caption{The computational complexity, processing latency, and output data size of each layer of different models.}
     \label{fig02}
 \end{figure*}

Third, non-uniformities in the DNN-based task processing pipeline are apparent. To elucidate this, we conducted a pilot study that analyzes execution latency by layer, intermediate output data size, and computational demands for Yolov5 \cite{nepal2022comparing}, AlexNet \cite{zhu2020multilevel}, and VGG16 \cite{mousavi2022novel}. The results, shown in Fig. \ref{fig02}, demonstrate that most of the computational cost is concentrated in the first half of the DNN, with considerable variability in latency, output data size, and computation across different layers of these DNNs. This underscores the need for uniform task partitioning. Arbitrary division can lead to significant imbalances in task assignment between UAVs, which may adversely affect the operational efficiency and survival time of multi-UAV, thus hindering its overall effectiveness in multi-UAV operations \cite{jouhari2021distributed}. To delineate the differences between the proposed DNN task assignment strategy and existing studies in this field, a comparison of various works is presented in Table \ref{tab:table01}. Performance metrics that meet the specified criteria are indicated with a $\checkmark$.

\subsection{Summary of Contributions}
In this paper, we develop a collaborative approach of multi-UAVs for the assignment of DNN tasks. This approach considers the computing resources, energy consumption, and number of layers they can execute each UAV. Moreover, our method introduces a finer-grained task partitioning compared to previous research. Our main contributions are as follows.
\begin{itemize}
\item{We propose a mother-and-child UAV system that employs a high-altitude platform (HAP), such as an airship, equipped with multiple UAV launchers. From this HAP, several UAVs are launched and networked to support collaborative tasks. Then we design DNN tasks directly on UAVs, performing inference independently of the ground base station. This approach enhances the scalability of the system. Furthermore, we propose a novel greedy algorithm for UAV path planning that utilizes a fitness function to minimize flight distance, taking into account the varying sizes of tasks.}

\item{We formulate the optimization problem of joint energy consumption, task completion rate, and load balancing in a multi-UAV network as a Mixed Integer Nonlinear Program. Subsequently, we introduce a multi-agent reinforcement learning (MARL) algorithm based on a generative diffusion model (GDM) to address the problem. Our work tackles constraints related to the number of UAVs, the AoI of the DNN task, the cache capacity, and the limitations of energy consumption. Notably, our research is the first to integrate a GDM into the MARL framework for multi-UAV networks.}

\item{The proposed algorithm leverages an innovative application of a GDM to determine optimal DNN task assignment decisions. We use the reverse denoising process of the GDM instead of the actor network of the multi-agent deep deterministic policy gradient (MADDPG) to generate specific DNN task assignment actions based on the agent's observations in a dynamic environment. The GDM functions by sequentially reducing noise through a series of denoising steps, thereby extracting optimal decisions from an initial state of Gaussian noise. By decomposing complex computations into simpler subtasks, we achieve substantial reductions in latency and improve the overall efficiency.}

%
\end{itemize}

The organization of this paper is as follows: Section \ref{sec:2} reviews the related work, the system model is formulated in Section \ref{sec:3}, Section \ref{sec:4} describes the proposed problem, Section \ref{sec:5} formulates the path planning approach, Section \ref{sec:6} describes the GDM-MADDPG approach, and Section \ref{sec:7} presents the simulation results, followed by the conclusion in Section \ref{sec:8}.

\section{Related Works}
\label{sec:2}

In this section, we summarize the contributions of the related works and discuss the distinctions between prior research and our methodology.

\subsection{Edge Intelligent for DNN Task}

The partition of the DNN task represents a significant research challenge that involves dividing a DNN into several segments and distributing these segments between designated locations. The existing work provides valuable information to support the efficient DNN inference task. The authors in \cite{gao2021task} present a joint design for task partitioning and offloading in DNN-enabled edge computing, involving a server and multiple devices. In \cite{li2024distributed}, the authors propose a multi-task learning-based asynchronous advantage actor-critic approach, and explore distributed DNN inference through fine-grained model partitioning in edge computing. These studies overlook the mobility of UAV networks. The authors in \cite{liu2023toward} explore accelerating DNN-based task processing in vehicular edge computing through task partitioning and offloading. In \cite{ren2024efficient}, the authors propose an efficient allocation strategy using RL and investigate fast DNN inference in UAV swarms through multi-UAV collaboration. However, studies have not addressed the adaptation challenges associated with the partitioning of DNN tasks. In addition, the oversight of these studies may result in decreased network stability in certain resource-constrained networks due to suboptimal task assignment within the DNN.


\subsection{MARL for Task Offloading}
MARL enables each agent to optimize its policy through observations of both the environment and the policies of other agents \cite{xu2021multi, liu2024ga}. Many existing MARL-based approaches significantly improve task offload efficiency within dynamic networks \cite{liu2019trajectory, liu2023energy}. The authors in \cite{kim2024cooperative} propose a dedicated actor neural network for coordination and a scalable training algorithm for the offloading of UAV-aided MEC tasks. In \cite{ju2023joint}, the authors present a DRL-based joint safe offloading and resource allocation scheme for vehicular edge computing networks. These studies have paid limited attention to this aspect. The authors in \cite{liu2024mobile} present a mobility-aware service offloading and migration scheme with Lyapunov optimization and an MADDPG algorithm. The authors in \cite{qin2023multi} focus on optimizing task offloading and UAV trajectory under energy and queue latency constraints. However, most of these studies focus on optimizing network performance metrics, including task completion latency and energy consumption. Our research emphasizes DNN inference services, where the AoI for DNN tasks is a critical performance metric. 


\subsection{GAI for UAV Networks}
GAI has the potential to effectively overcome the limitations of conventional AI and can be applied to optimize UAV networks, particularly to improve transmission rates, communication capacities, and energy efficiency \cite{sun2024generative, gao2024guiding}. In \cite{zhang2023rme}, the authors present a radio propagation model with a conditional generative adversarial network. The authors in \cite{van2024generative} introduce a two-stage generative neural network to predict link states and generate path losses, latencys, and arrival and departure angles in millimeter wave communication for UAVs. The authors in \cite{liu2023joint} explore the long-term optimization of joint task offloading and resource allocation in a multi-access edge computing network for UAVs. In \cite{kang2023cooperative}, the authors present a hierarchical aerial computing system where UAVs collect tasks from ground devices and offload them to HAPs. Unlike the above schemes, our approach incorporates a path planning, GDM-based MARL decision-making framework specifically tailored for multi-UAV networks. This methodology aims to minimize the cost associated with DNN task completion while ensuring the system's stability.

\section{System Model}
\label{sec:3}
In this section, we first present a novel network architecture, then introduce in detail the DNN task model, the communication model, the mobile model, the AoI model, and the energy consumption model, respectively.

\subsection{Network Architecture}
\begin{figure}[!t]
	\centering
	\includegraphics[width=3.5in]{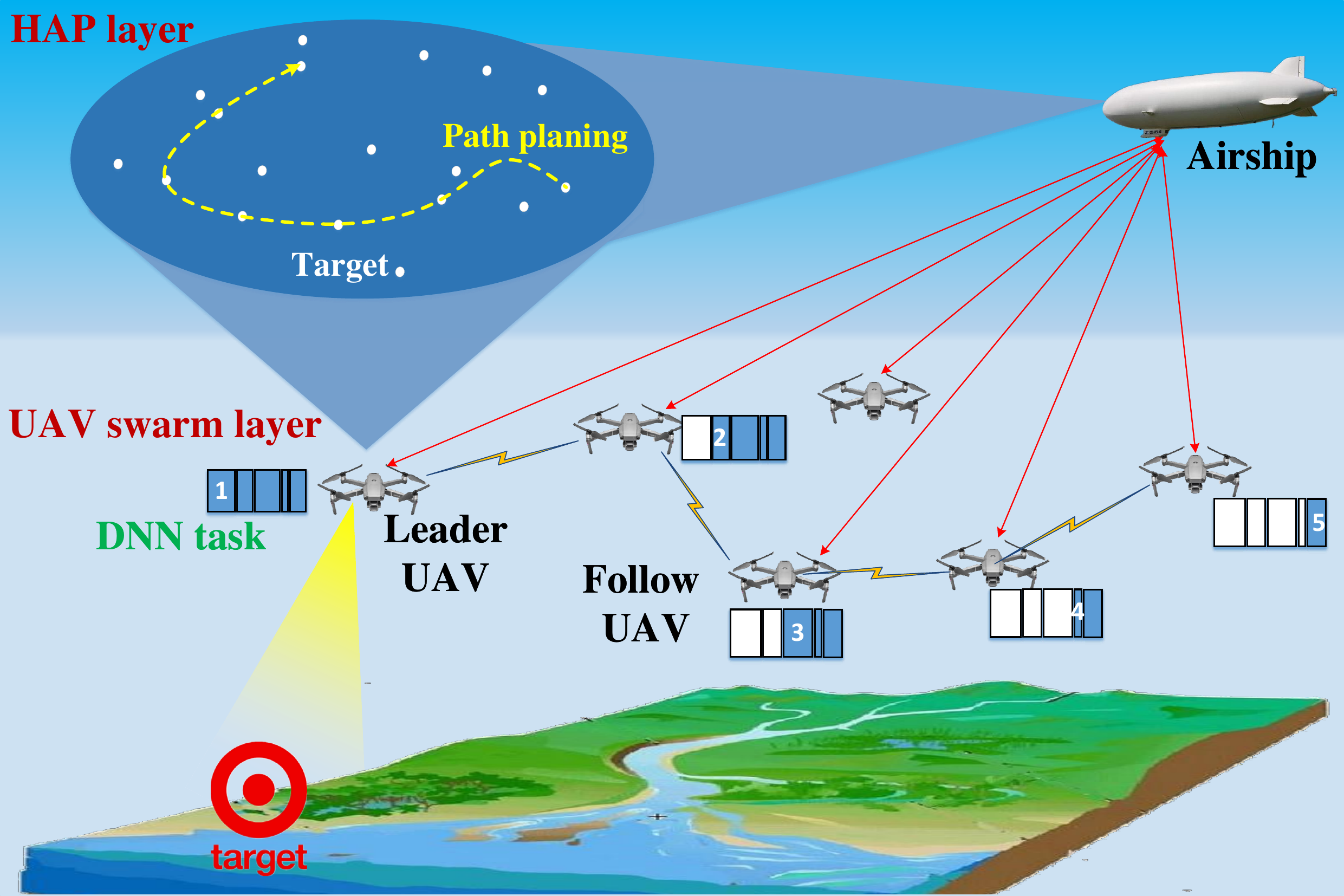}%
	\caption{Network architecture.}
	\label{fig03}
\end{figure}

The network architecture is primarily composed of an HAP layer and a UAV swarm layer, as illustrated in Fig. \ref{fig03}. The HAP layer consists of a tethered platform with extensive computing and communication capabilities, such as an airship. This platform exchanges information with UAVs via wireless links to facilitate model training \cite{tang2023digital} and path planning (see Section \ref{sec:5}). The airship will transmit the final result of the path planning to the leader UAV. The UAV swarm comprises a leader UAV and multiple follow UAVs. All UAVs are isomorphic, with the total number of UAVs being ${n} = \left\{ {1,2, \cdot  \cdot  \cdot, N} \right\}$, and each is equipped with trained DNN models. The leader UAV receives the results of the path planning transmitted by the airship and operates according to the predetermined route. It is also responsible for collecting and processing images and point cloud data of the target area, including tasks such as target detection and mapping. Thus, it functions as both a task producer and executor, as well as a decision maker for task assignment. The follow UAVs primarily execute the tasks assigned by the leader UAV, with the assignment of DNN tasks between follow UAVs being streamlined. The leader UAV evaluates the task size and the remaining resource status of the follow UAVs to determine whether to assign tasks. Therefore, to prevent task overload on the UAVs, extend the operational lifespan of the UAV swarm, and improve the timeliness of task processing and network stability, efficient task assignment decisions for the UAV swarm are particularly crucial. 

\subsection{DNN Task Model}
The area where the task is performed in this paper consists of $q$ center coordinates of the target area, denoted as $q=1, 2, ... , \mathbb{W}$, and the sequence of the UAV inspection target coordinates is represented as $\mathfrak{q}(q)=[\mathfrak{q}(1),\ldots,\mathfrak{q}(\mathbb{W})]$. The data size in the target area is
$\ell_q=\{1,...,\mathbb{L}_\mathbb{W}\}$. Assume that the number of DNN tasks in each synchronization cycle is $i=\{1,2,...,I\}$, the DNN type is $\kappa=\{1,...,\mathbb{K}\}$. The DNN model typically comprises $l$ layers, denoted as $l=\{1,2,...,L\}$. Each layer has a cache capacity $m_{i,l}$ and a computing requirement ${c}_{i,l}$, represented by a tuple $(m_{i,l},c_{i,l})$. $l_\kappa=\{1,2,...,L_\kappa\}$ represents a set of layers associated with a specific type of DNN task, including convolutional and pooling layers for feature extraction, as well as fully connected layers for classification \cite{liu2024dnn}. The DNN task is decomposed into multiple subtasks for serial processing, and each follow UAV serves as an execution stage in this serial processing. This approach allows each subtask to offload and transfer intermediate results among different follow UAVs without requiring the processing of the entire task, thereby enhancing the computational efficiency and real-time performance of computationally intensive and latency-sensitive tasks \cite{qu2023elastic}.

In response to such practical needs, determining the optimal split point of the task is crucial in DNN task assignment. Assume that the split point of the ${i}$-th DNN task model is $P_{i,n^{\prime}}$, and $n^{\prime}\in\{1,2,...,N-1\}$ represents the number of subtasks generated after the DNN task model is divided into blocks, its size depending on the initial task data size. Based on the size of the task data, we can categorize the task assignment modes into swarm assignment, partial assignment, and binary assignment. Swarm assignment indicates that the DNN task is processed serially through pipeline assignment across all follow UAVs. Partial assignment means that the DNN task is processed serially through pipeline assignment among some follow UAVs. Binary assignment entails that the DNN task is completely offloaded to a single follow UAV for processing. The DNN task assignment mode ${M}$ is expressed as follows
\begin{equation}
\label{E22}
M=f(k,\ell,\tau_i^{max},O),
\end{equation}
where $f(\cdot)$ represents the function that determines the task assignment mode (see Section \ref{sec:6}). $\tau_i^{max}$ denotes the maximum latency allowed for the total completion time of each task, and $O$ represents the Observation space of the follow UAV.

\subsection{Communication Model}
DNN task assignment is performed within the UAV swarm, and during the execution of the task, the UAV swarm maintains a fixed formation flight. At time slot $t$, the coordinates of UAV are designated as $\mathcal{P}^{n}=[x^n_t,y^n_t,z^n_t]$, and the distance between the two UAVs is defined as follows
\begin{equation}
\label{E21}
d^{n,n+1}_t=\sqrt{\left\|\mathcal{P}^n-\mathcal{P}^{n+1}\right\|^2}.
\end{equation}

The UAV-to-UAV data link utilizes line-of-sight communication, and its channel model is described by the Close-In Free Space Reference Model (CI) \cite{liu2019trajectory, li2024gan}. The path loss is expressed as follows
\begin{equation}
\label{E20}
\begin{aligned}
PL_{CI}(d^{n,n+1}_t,\frak{f})=PL_{{FS},ref}(\frak{f})\\+10n_{CI}\mathrm{log}_{10}\left(d^{n,n+1}_t\right)+\xi_{\sigma,CI},
\end{aligned}
\end{equation}
where $PL_{{FS},ref}(\frak{f})$ represents the free space path loss per unit length. $\frak{f}$ means the frequency of a radio wave is equal to the speed of light divided by the wavelength. $n_{CI}$ is the path loss index, and $\xi_{\sigma,CI}$ is the shadow fading parameter, which is typically assumed to follow a zero-mean Gaussian distribution.

The signal-to-noise ratio between the two UAVs is formulated as follows
\begin{equation}
\label{E19}
SINR^{n,n+1}_t=\frac{P_n-PL_{CI}(d^{n,n+1}_t,\frak{f})}{P_I+P_N},
\end{equation}
where $P_{n}$ is the information transmission power of the UAV. $P_{I}$ is the interference power.  $P_{N}$ is the noise power, denoted as $P_{N}=kBT$, where ${k}$ is the Boltzmann constant. ${B_n}$ is the channel bandwidth. ${T}$ is the absolute temperature.

Therefore, the data transmission rate between the UAVs can be expressed as follows
\begin{equation}
\label{E18}
r^{n,n+1}_t=B_n\log_2\left(1+SINR^{n,n+1}_t\right).
\end{equation}

\subsection{Mobile Model}
Let $\boldsymbol{\varphi}_{q}$ denote the binary decision variable for the coordinate of the inspection target $\boldsymbol{g}_{\mathfrak{q}(q)}$; specifically, when the target coordinate $\boldsymbol{g}_{\mathfrak{q}(q)}$ is inspected by the UAV, $\boldsymbol{\varphi}_{q}$ is equal to 1. Otherwise, $\boldsymbol{\varphi}_{q}$ is equal to 0. After the UAV swarm collects data from the target coordinate $\boldsymbol{g}_{\mathfrak{q}(q)}$, it must process the data before reaching the next target coordinate $\boldsymbol{g}_{\mathfrak{q}(q+1)}$. The Euclidean distance between the two target coordinates is denoted as $D^{\mathfrak{q}(q),\mathfrak{q}(q+1)}$. The total moving distance of the UAV to complete the inspection task is the sum of the distances of each segment along its flight path, as illustrated below
\begin{equation}
\label{E180}
D^{total}=\sum_{\mathfrak{q}(1)}^{\mathfrak{q}(q)}D^{\mathfrak{q}(q),\mathfrak{q}(q+1)}.
\end{equation}

Assuming that the UAV flies at a constant speed $\nu$, the time taken by the UAV swarm to travel from one coordinate to the next one is expressed as follows
\begin{equation}
\label{E17}
t^{next}=D^{\mathfrak{q}(q),\mathfrak{q}(q+1)}/\nu.
\end{equation}

The total flight time for the process of the UAV executing the task is represented as follows
\begin{equation}
\label{E16}
t^{fly}=D^{total}/\nu.
\end{equation}

\subsection{AoI Model}
AoI defined in this paper aims to measure the duration from data collection to processing completion, incorporating waiting latency, transmission latency, and computation latency to reflect the real-time performance of DNN task assignment. We assume that each UAV has a maximum memory capacity $m_n^{threshold}$, available energy $e_n^{threshold}$, and computing capacity $f_{n}$. The AoI is optimized by minimizing the UAV's moving distance and refining the DNN task assignment strategy. The waiting latency for the data collected by the leader UAV to be processed by the DNN model is denoted as
\begin{equation}
\label{E15}
t_{i,n}^{await}=T_{i^{\prime}}-T_i,
\end{equation}
where $T_{i}$ represents the moment when the collected raw data is obtained on the leader UAV or when the subtask ${n}$ is generated on a follow UAV. $T_{i^{\prime}}$ signifies the moment when the collected raw data is processed in the leader UAV or when the subtask ${i}$ is about to be sent from the follow UAV ${n}$ to the follow ${n+1}$.

The latency for the intermediate result of the completed DNN task transmitted by follow UAV ${n}$ is expressed as
\begin{equation}
\label{E13}
t_{i,n}^{trans}=\frac{W_{i,l}}{r^{n,n+1}_t}, 
\end{equation}
where $W_{i,l}$ is the data size output by the previous follow UAV, specifically the data size before the split point 
$p_{i,n^{\prime}+1}$.

The computation latency of the task is represented as
\begin{equation}
\label{E12}
t_{i,n}^{comp}=\sum_{l=p_{i,n^{\prime}-1}+1}^{p_{i,n^{\prime}}}\frac{c_{i,l}}{f_n},
\end{equation}
where $\frac{c_{i,l}}{f_{n}}$ denotes the computing time of the ${i}$-th subtask. $c_{i,l}$ indicates the computing requirement of the ${i}$-th subtask, and $f_{n}$ represents the computing capacity of the $n$-th UAV.

Therefore, the latency $\mathcal{T}$ for the complete DNN task and the execution latency of the entire DNN task ${AoI}$ are as follows
\begin{equation}
\label{E11}
\mathcal{T}=\sum_{i=1}^{I}\sum_{n=1}^{N}t_{i,n}^{trans}+t_{i,n}^{comp},
\end{equation}
\begin{equation}
\label{E10}
AoI=\sum_{i=1}^{I}\sum_{n=1}^{N}t_{i,n}^{{await}}+t_{i,n}^{trans}+t_{i,n}^{comp}.
\end{equation}

\subsection{Energy Consumption Model}
The energy consumption of UAVs in performing DNN tasks primarily consists of computing, transmission, and flight energy consumption.

1) The computing energy consumption of the UAV is denoted as
\begin{equation}
\label{E09}
e_n^{comp}=\sum_{i=1}^I\sum_{l=p_{i,n^{\prime}-1}+1}^{p_{i,n^{\prime}}}k_0f_n^2{c_{i,l}},
\end{equation}
where $k_{0}$ represents the energy efficiency parameter and $c_{i,l}$ denotes the computing requirement or complexity of the ${l}$-th layer in the ${i}$-th task.

2) The transmission energy consumption is approximated as the product of the transmission power $P_{n}$ and the transmission time $t_{i,n}^{trans}$ of the intermediate result of the DNN task as follows
\begin{equation}
\label{E08}
e_n^{trans}=\sum_{i=1}^IP_nt_{i,n}^{trans}.
\end{equation}

3) As one of the energy consumptions that must be considered, the flight energy consumption of the UAV is adopted in \cite{liu2023energy}. The flight energy consumption is approximated as the product $e_n^{fly}$ of the propulsion power $P^{fly}$ of the UAV and the total flight time during the task execution as follows
\begin{equation}
\label{E07}
e_n^{fly}=P^{fly}t^{fly},
\end{equation}
\begin{equation}
\label{E06}
\begin{aligned}P_n^{fly}&=P_1\left(1+\frac{3\nu^2}{\nu_{tip}^2}\right)+P_2\left(\sqrt{1+\frac{\nu^4}{4\nu_0^4}}-\frac{\nu^2}{2\nu_0^2}\right)\\&+\frac12\zeta\rho\varsigma{s}\nu^3,\end{aligned}
\end{equation}
where $P_{1}$ represents the blade shape power in the hovering state, $P_{2}$ is the induced power, $\nu_{tip}$ is the speed of the rotor blade tip, and $\nu_{0}$ is the average induced rotor speed in the hovering state. Additionally, $\zeta$, $\rho $, $\varsigma$, and $s$ refer to the fuselage drag ratio, air density, solidity of the rotor, and disk area, respectively. Under hovering conditions, the power consumption of the UAV is the sum of $P_{1}$ and $P_{2}$.

Consequently, the energy consumption of a follow UAV is $e_n=e_n^{comp}+e_n^{trans}+e_n^{fly}$. And then the total energy consumption of all DNN tasks executed by the UAV swarm is denoted as
\begin{equation}
\label{E05}
E=\sum_{n=1}^Ne_n^{comp}+e_n^{trans}+e_n^{fly}.
\end{equation}

\section{Problem Formulation}
\label{sec:4}
In this paper, AoI and load balancing are fully considered in the DNN task assignment of UAV swarms. Generally, completing tasks solely by UAVs is a common method to enhance AoI. However, this approach often leads to a significant increase in the energy consumption of individual UAVs, thus reducing the overall survival time of the UAV swarm. To optimize load balancing in task assignment, frequent DNN task transfers within the UAV swarm may also extend task execution time and potentially deteriorate the timeliness of the collected data. Therefore, during the execution process of the task, it is essential to establish a reasonable partitioning of the DNN tasks and an assignment scheme that effectively balances the task completion rate $\eta=l_{\kappa}^{completed}/l_{\kappa}$, AoI and load balancing to improve the stability of the UAV swarm.

To this end, this paper defines a utility function \( U \) for the assignment of DNN tasks, which includes three components: the utility \( u_1 \), representing the contribution of a single UAV to a specific task; the utility \( u_2 \), which reflects the completion rate of the task and emphasizes the importance of both completing the task and improving AoI; and the UAV load balance utility \( u_3 \), which indicates the variance of the remaining energy of the UAVs. This last component is crucial for maintaining balanced energy consumption across the UAV swarm.

\begin{equation}
\label{E04}
U=\delta{u}_1+\varepsilon{u}_2+\theta u_3,
\end{equation}

\begin{equation}
\label{E03}
{u}_1=e_n^{comp}+e_n^{trans},
\end{equation}

\begin{equation}
\label{E02}
{u}_2=\begin{cases}\alpha\eta_\kappa-\beta{AoI}_{i}, \text{task is completed}\\\gamma(l_{\kappa}^{completed})-\beta\tau_{i}^{max}, \text{otherwise}\end{cases},
\end{equation}

\begin{equation}
\label{E00}
\begin{aligned}{u}_{3}&=\sum_{n=1}^N\left\{e_n^{threshold}-e_n\right.-\left[\sum_{n=1}^Ne_n^{threshold}-e_n\right]/N\}^2\end{aligned},
\end{equation}
where $\alpha,\beta,\gamma,\delta,\varepsilon,\theta$ are weight coefficients, $l_{\kappa}^{completed}$ indicates the number of layers of the $\kappa$-th DNN model that have been completed. 

Taking into account the constraints related to the number of UAVs, cache capacity, energy consumption, and latency, we reformulate the optimization problem of assigning DNN tasks by determining the split point of the task $p_{i,n}$ and the number of UAVs of wing men ${n}$ that correspond to the task. The optimization objective function $\mathbb{P}_{1}$ is defined as follows 
\begin{equation}
\label{E000}
\underset{n, p_{i,n^{\prime}}}{\max} U
\end{equation}
\begin{align}
\textbf{s.t.}
&C_1:\quad 1 \leq p_{i,1} < p_{i,2} < \cdots < p_{i,N-1} \leq L-1 \tag{23a} \\
&C_2:\quad 1 \leq n \leq N \tag{23b} \\
&C_3:\quad \sum_{l=p_{i,n^{\prime}-1}+1}^{p_{i,n^{\prime}}} m_{i,l} \leq m_n^{{threshold}} \tag{23c} \\
&C_4:\quad e_n^{{comp}} + e_n^{{trans}} + e_n^{{fly}} \leq e_n^{{threshold}} \tag{23d} \\
&C_5:\quad e_n^{{rendez}} \leq e_n^{{threshold}} - e_n \tag{23e} \\
&C_6:\quad \sum_{q=1}^\mathbb{W} \varphi_q = 1 \tag{23f} \\
&C_7:\quad t_{i,n}^{{await}} + t_{i,n}^{{trans}} + t_{i,n}^{{comp}} \leq \tau_i \tag{23g}
\end{align}
where $C_{1}$ signifies that the DNN model split point is located within the hierarchy. $C_{2}$ indicates that the number of unloaded tasks falls within the total number of UAVs. $C_{3}$ specifies that the task size does not exceed the maximum cache capacity of the UAV. $C_{4}$ ensures that the total energy consumed does not exceed the UAV's energy capacity. $C_{5}$ denotes that the remaining energy consumption of the UAV is greater than the energy consumption $e_n^{rendez}$ required to return. $C_{6}$ requires that the leader UAV starts from the initial coordinate and visits at least one target coordinate. $C_{7}$ stipulates that each task must not exceed the maximum allowable latency.

As shown in the previous section, the objective optimization function $\mathbb{P}{1}$ is an NP-hard problem that involves both discrete and continuous variables. The problem $\mathbb{P}{1}$ is divided into a path planning subproblem and an MDP subproblem, and then solved using the proposed GDM-MADDPG with a path planning approach.

\section{Path planning method based on greedy algorithm}
\label{sec:5}
The greedy algorithm is used to optimize the path of the UAV due to its low computational cost and fast convergence \cite{tian2023uav}. A stochastic component is introduced to the greedy algorithm, which considers the task size within the target area, thereby providing an approximate solution for the globally optimal UAV flight path.

\subsection{Fitness Function}
This paper employs a greedy algorithm to minimize the total flight distance of the UAV while jointly considering the task size of the target area to address the UAV path planning problem. Before executing the DNN task assignment, the UAV must plan an optimal flight path to reduce task completion time and energy consumption. Given that the task sizes of each target area vary, we denote the task size of the coordinate of the $q$ -th target as $\mathbb{L}_{q}$, and let ${\upsilon}_{q}$ represent the average rate at which the UAV swarm processes the tasks. Consequently, the time required to process the task size at the $q$-th target coordinate is denoted as
\begin{equation}
\label{E001}
t_{q}=\mathbb{L}_{q}/\upsilon_{q}.
\end{equation}

To improve task processing efficiency and AoI, the UAV swarm is required to complete the task after inspecting the target coordinate and before reaching the next target coordinate, thus establishing the condition $t_{q}\leq t^{next}$. Furthermore, we define $\Delta t=t^{next}-t_{q}$ as the difference between flight time and task processing time between two target coordinates of the task.

Accordingly, the fitness function integrates the UAV flight distance and the quantity of tasks processed by the DNN model, represented as
\begin{equation}
\label{E002}
F=\vartheta D^{total}+\rho\Delta t,
\end{equation}
where $\vartheta$ and $\rho$ are weight coefficients utilized to balance the trade-off between the UAV flight distance and the amount of data processed by the DNN model.

\begin{algorithm}
    \SetAlgoLined 
	\caption{Path planning method based on greedy algorithm}\label{alg1}
	\KwIn{Coordinates of targets, task size of target coordinates $\ell_q$}
	\KwOut{The sequence of the UAV inspection target coordinates}
        Initial parameters of path planning, such as $\widetilde{\mathbb{W}}$, $\vartheta$, $\rho$, $k$, $v$, ${f}_{q}$  
        
        Execute fitness function as Eq. \eqref{E002} 
        
        Select the first target coordinate \[\mathfrak{q}(1)=\arg\max_{\mathfrak{q}(q)\in\widetilde{\mathbb{W}}}\mathbb{L}_\mathfrak{q} \!\left/ \!\right. \! D_{\mathfrak{q}(q), \mathfrak{q}(0)}\] 

        
	\For{${i=1,2,...,q}$}{
		\eIf{$|\widetilde{\mathbb{W}}|\leq k$}{
			$\tilde{\mathrm{g}}=\widetilde{\mathbb{W}}$
		}{
			$\tilde{\mathrm{g}}\subset\widetilde{\mathbb{W}}$
		}
	}
\end{algorithm}

\subsection{Path Planning Method}
The greedy algorithm is a straightforward, lightweight heuristic solution method that is easy to implement \cite{tian2023uav}. However, this approach can lead to suboptimal outcomes, as selecting local optimal solutions does not guarantee a global optimal solution. To address this limitation, this paper incorporates randomness into the design of the greedy algorithm to mitigate the risk of converging on a local optimal solution.

Specifically, the candidate target coordinates $k$ are randomly selected from the set of uninspected target coordinates, denoted $\widetilde{\mathbb{W}}$. If $|\widetilde{\mathbb{W}}|\leq\mathrm{k}$ indicates that all uninspected target coordinates are chosen as candidate target coordinates $k$, then we define this set as $\tilde{\mathrm{g}}=\widetilde{\mathrm{W}}$. Otherwise, $\widetilde{\mathbb{W}}$ uninspected target coordinates $k$ are randomly selected from $\tilde{\mathrm{g}}\subset\widetilde{\mathbb{W}}$. For each candidate target coordinate, the fitness function value ${F}$ is calculated and the target coordinate with the smallest value $F$ is selected as the next target coordinate. A lower value of $F$ corresponds to a lower total cost. The chosen target coordinate is then added to the flight path sequence and removed from the set of candidate target coordinates. This process is repeated until all target coordinates have been inspected once, at which point the UAV returns to the initial target coordinate.

The path planning algorithm based on the greedy algorithm is illustrated in Algorithm \ref{alg1}. In this context, the path planning problem is denoted as $\mathbb{P}_{2}$

\begin{equation}
\label{E003}
{min~}F
\end{equation}
\begin{align}
\textbf{s.t.}
&C_8:\mathrm{t_q\leq t^{next}} \tag{26a}
\end{align}
where $\mathrm{C}_{8}$ signifies the requirement that the UAV swarm completes the task at the previous target coordinate before proceeding to the next target coordinate.

\begin{figure*}[htb]
    \centering
	\includegraphics[width=\textwidth]{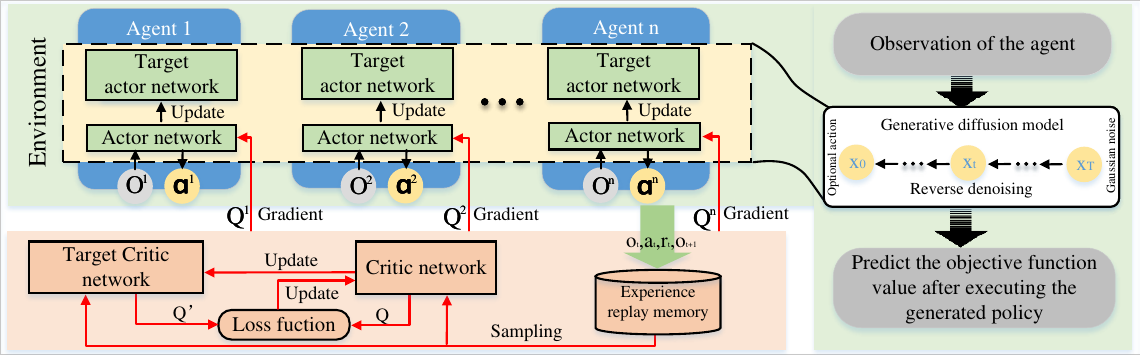}%
	\caption{GDM-MADDPG for strategy generation.}
	\label{fig04}
\end{figure*}

\section{DNN task assignment algorithm based on GDM-MADDPG}
\label{sec:6}
GDM demonstrates robust generative capabilities and is adept at navigating complex dynamic decision optimization scenarios. Notably, it can identify optimal solutions even in the absence of a dedicated dataset. This section focuses on the design methodology and algorithmic framework of the GDM-MADDPG method. It also encompasses the design and modeling of both the denoising process and MDP.

\subsection{Design of GDM-MADDPG Algorithm}


DNN Task assignment decision generation differs from traditional applications of GDM, primarily because decision optimization often lacks a large, dedicated dataset for offline training. In this paper, the DNN task assignment decision process involves the GDM predicting noise distribution through iterative refinement, followed by training its reverse denoising process, specifically denoising Gaussian noise to yield optimal action decisions. This approach transforms the challenge of limited datasets into an opportunity for dynamic online learning and agent self-optimization. GDM-MADDPG takes the observation of the agent as the condition of the denoising network of the diffusion model, and replaces the actor network of MADDPG with the reverse denoising process to better capture the correlation between the data, as shown in Fig. \ref{fig04}. In the initialization phase, each agent will establish its own Actor network and Critic network, as well as the corresponding target network, and create an experience replay buffer to store interaction data. The Actor network determines the action based on the local observation output of the agent, while the Critic network evaluates the expected reward of the state-action pairs of all agents. During training, the agent performs actions and collects environmental feedback, which is stored in the experience replay memory. In order to stabilize the training, GDM-MADDPG introduces a target network whose parameters are updated at a certain rate to achieve a smooth learning process. Next, samples are randomly drawn from the memory to update the Critic network so that it can more accurately predict the rewards of Observation-action pairs. Subsequently, the output of the Critic network is used to update the Actor network with the goal of maximizing the expected reward.


\subsection{Reverse Denoising Process of GDM}
The GDM inverse denoising process serves as a reverse denoising technique that reconstructs the original data from noisy observations. In this context, the denoising network, once fully trained, is capable of generating optimal decisions for DNN task assignment in any dynamic environment. Both the selection of assignment modes and the design of utility functions typically adapt to the dynamic variations present in the environment. This capability to respond to changing conditions and generate appropriate actions is invaluable in decision optimization. Consequently, the GDM inverse denoising process can leverage information such as the UAV observation space ${O}$, task data size $\ell_{q}$, and task type $\kappa$ as conditions within the denoising network, thus facilitating adaptive decision generation.

For the agent ${n}$, the objective of the inverse process is to infer the probability distribution $\mathbf{x}_0^n$ of each task that is selected from the Gaussian noise $\mathbf{x}_T^n{\sim}N(0,I)$. If the inverse distribution $p(\mathbf{x}_{t-1}|\mathbf{x}_t)$can be learned, it becomes possible to sample $\mathbf{x}_{t}$ from the standard normal distribution ${N}(0,{I})$ and subsequently obtain samples from $p(\mathbf{x}_0)$ through the inverse process. However, statistical estimation $p(\mathbf{x}_{t-1}|\mathbf{x}_t)$ requires calculations that involve the complexity of the data distribution, which is often challenging to manage in practice. Therefore, our goal is to estimate $p(\mathbf{x}_{t-1}|\mathbf{x}_t)$ using the parameterized model $p_{\theta}$ as outlined below
\begin{equation}
\label{E0098}
p_\theta(\mathbf{x}_{t-1}|\mathbf{x}_t)=\mathcal{N}(\mathbf{x}_{t-1};\boldsymbol{\mu}_\theta(\mathbf{x}_t,t),\boldsymbol{\Sigma}_\theta(\mathbf{x}_t,t)).
\end{equation}

In this manner, we can derive the trajectory from $\mathbf{X}_{T}$ to $\mathbf{x}_0$ as follows

\begin{equation}
\label{E0088}
p_\theta(\mathbf{x}_{0:T})=p_\theta(\mathbf{x}_T)\prod_{t=1}^Tp_\theta(\mathbf{x}_{t-1}|\mathbf{x}_t).
\end{equation}

By conditioning the model in the time interval ${t}$, it can learn to predict the parameters of the Gaussian distribution, specifically the mean $\mu_\theta(\mathbf{x}_t,t)$ and the covariance matrix $\boldsymbol{\Sigma}_\theta(\mathbf{x}_t,t)$ at each time step.

The training of GDM involves optimizing the negative log-likelihood of the training data. According to the literature \cite{ho2020denoising}, by incorporating conditional information $g$ in the denoising process, it can be modeled as a noise prediction model with the covariance matrix fixed to $p_\theta(\mathbf{x}_{0:T})$
\begin{equation}
\label{E0038}
\Sigma_\theta(\mathbf{x}_t,g,{t})=\beta_{t}{I}.
\end{equation}

The construction of the mean is as follows
\begin{equation}
\label{E0039}
\mu_\theta(x_t,g,t)=\frac1{\sqrt{\alpha_t}}\Bigg(x_t-\frac{\beta_t}{\sqrt{1-\overline{\alpha}_t}}\epsilon_\theta(x_t,g,t)\Bigg).
\end{equation}

We first sample $\mathbf{x}_{T}$ from ${N}(0,{I})$ and then sample $x_{t-1}|x_t=\frac{x_t}{\sqrt{\alpha_t}}-\frac{\beta_t}{\sqrt{\alpha_t(1-\overline{\alpha}_t)}}\epsilon_\theta(x_t,g,t)+\sqrt{\beta_t}\epsilon $ through the inverse denoising process parameterized by $\theta $
where $\epsilon\sim{N}(0,{I})$ and \( t = 1, \ldots, T \).
The loss function for the model training of the denoising network is given by
\begin{equation}
\label{E01}
\mathcal{L}_t=\mathbb{E}_{\mathbf{x}_0,t,\boldsymbol{\epsilon}}[\|\boldsymbol{\epsilon}-\boldsymbol{\epsilon}_\theta(\sqrt{\overline{\alpha}_t}\mathbf{x}_0+\sqrt{1-\overline{\alpha}_t}\boldsymbol{\epsilon},t)\|^2].
\end{equation}

\subsection{MDP Modeling}
GDM effectively addresses challenges within the DNN task pipeline assignment environment, generating optimal assignment decisions that maximize the utility function. This method showcases the ability of GDM to manage complex data distributions and produce high-quality decisions, particularly in dynamic and evolving environments \cite{liu2024dnn}. By integrating DRL, more efficient and adaptable action strategies can be developed. Consequently, the utility maximization problem is formulated as a multi-agent Markov decision process (MDP), represented by a tuple $(n,O,\mathcal{A},{r},\gamma)$, 
where $\text{n}$ denotes the number of agents, ${O}$ is the observation space of the agent, $\gamma$ represents the discount factor, and ${r}$ signifies the reward obtained by the agent. The agent is represented by the UAV. It learns an optimal task assignment strategy based on GDM-MADDPG according to task size, computing resources, and task latency to achieve optimal load balancing and AoI. The DNN task assignment decision generation based on GDM-MADDPG is detailed in Fig. \ref{fig04} and Algorithm \ref{alg2}.

\begin{algorithm}
    \SetAlgoLined 
    \caption{GDM-MADDPG Algorithm}\label{alg2}
    
    Initialize the parameters of the Critic network and the Actor network (denoising network)

    \For{$\text{episode}=0\to \varepsilon$}{
        Reset the environment and the initial observation $O$

        \For{$t=1,2,\ldots,\mathbb{T}$}{

            \For{$n=1, 2, \ldots, N$}{
                Use the observation space of agent $n$ as a condition of denoising network to generate adapted actions
                
                \For{$t=1, 2, \ldots, T$}{
                    Predict the noise $\varepsilon$

                    Gaussian noise $\mathbf{x}_{T}$ is denoised in reverse process, obtaining $\mathbf{x}_{0}^{n}$

                    Agent $n$ selects an action $\mathbf{x}_{0}^{n}$ according to $a_{t}^{i}=f(\mathbf{x}_{0}^{i})$

                    Execute the chosen action $a_{t}^{i}$

                    Obtain the reward $\mathbf{x}_{0}^{n}$ and the next observation $o_{t}^{n^{\prime}}$

                    $o_t^n\leftarrow o_t^{n^{\prime}}$

                    \eIf{the buffer is not full}{
			             Store the training data $((O_{t},A_{t},R_{t},O_{t}^{\prime})$ into the buffer
		              }{
			             Update the experience replay buffer
		              }  
                }
                \For{$n=1, 2, \ldots, N$}{
                    $b$ samples are taken from the buffer
                    $(O_{j}, A_{j}, R_{j}, O_{j}^{\prime}), \forall{j}=1, 2, \ldots, T$

                    Calculate the target value

                    Minimize the loss function to update the parameters of the Critic network
                    
                    Update the Actor and Critic target network parameters for each agent according to Eq. \eqref{E01}
                }
            }
        }
    }
\end{algorithm}

\textbf{Observation Space:} In each time slot $\text{t}$, the UAV collects observations from its environment. The observation space is denoted as $O_t=\{o_t^1, o_t^2, \ldots, o_t^N\}$, with the observation of a single agent represented by
\begin{equation}
\label{E0114}
o_t^n=\{\ell_t, \kappa_t,l_{\kappa,t}^{\mathrm{completed}},\tau_t, \mathbf{e}_t,\mathbf{m}_t, \boldsymbol{\mathcal{P}}_t,\mathfrak{q}(q)\},
\end{equation}
where $\ell_{t}$ indicates the amount of data associated with the DNN task at time slot ${t}$, $\kappa_t$ represents the type of DNN model, $l_{\kappa, t}^{completed}=\left\{l_1^{completed}, l_2^{completed}, \ldots, l_l^{completed}\right\}$ indicates the degree of completion of the DNN task (i.e., the number of completed layers), $\mathbf{\tau}_i^{max}=\{\tau_1^{max}, \tau_2^{max}, \ldots, \tau_I^{max}\}$ denotes the maximum latency that each task can tolerate at time slot ${t}$, $\mathbf{e}_t$ represents the energy consumption constraint of the UAV (the remaining energy at time slot ${t}$, $\mathbf{m}_t$ indicates the memory constraint of the UAV (the remaining memory at time slot ${t}$), $\boldsymbol{\mathcal{P}}_t=\{\mathcal{P}_t^1,\mathcal{P}_t^2, \ldots, \mathcal{P}_t^n\}$ represents the positions of all UAVs at time slot ${t}$, and $\mathfrak{q}_t(q)$ signifies the positions of all targets.

\textbf{Action Space:} The action space is defined as

$\mathcal{A}^n=\{a_1^n,a_2^n,\ldots a_t^n\}$. In each time slot $\text{t}$, the agent's action is denoted by $a_t^i$, where $i=\{1,2,...,I\}$ represents the task chosen by the agent for execution and $a_t^n=\{a_t^1,a_t^2,...,a_t^i\}$ means the action taken.

\textbf{Reward Function:} The agent's actions are constrained by AoI and are closely related to load balancing and task completion rates. To maximize the system utility, the agent observes $o_t^n$ and executes action $a_t^i$ to obtain the reward $r_t^n$ as follows
\begin{equation}
\label{E0014}
r_t^{n,{indiv}}=\begin{cases}\quad\delta{u}_1, e_n^{threshold}-e_n\geq e^{rendez}\\-\sigma(e_n^{threshold}-e_n), otherwise\end{cases},
\end{equation}

\begin{equation}
\label{E0024}
    r_t^{n,{group}}=\varepsilon u_2+\theta u_3,
\end{equation}

\begin{equation}
\label{E0034}    
    r_t^n=\vartheta r_t^{n,{indiv}}+(1-\vartheta)r_t^{n,{group}},
\end{equation}
where $r_t^{n,{indiv}}$ and $r_t^{n,{group}}$ represent individual and global rewards, respectively. $\sigma$ and ${\vartheta}$ are weight coefficients, and $e^{rendez}$ denotes the energy consumption required for the UAV to return.

\section{Experiment Setup and Performance Evaluation}
\label{sec:7}
To assess the proposed DNN task assignment approach, we conduct numerical experiments. The analysis includes a comparison with several benchmark algorithms to demonstrate its effectiveness and advantages.

\subsection{Experiment Setup}

\textit{1) Training Environment}

The hardware configuration consists of an Intel i7-13700K CPU and an NVIDIA RTX 4090 with 24 GB of RAM. The software environment includes pytorch GPU version 1.11.0 and Python version 3.8. The parameters for DNN are specified as follows: for the Actor network, the number of neurons in the first and second hidden layers are set to 256, respectively, while the number of neurons in the third hidden layer is determined by the dimensionality of possible actions. For the Critic network, the neuron counts for the three hidden layers are set to 256, respectively. During training, the learning rates for the Actor and Critic networks are set to \( \gamma = 10^{-3} \). The mini-batch size during training is 512. The discount factor is set to \( \gamma_m = 0.9 \), and the initial value and the decay rate for exploration are defined as \( \epsilon_0 = 0.9 \) and \( \beta = 10^{-4} \), respectively. In our simulation, we examine the performance of collaborative DNN task assignment in different data sizes. To simplify calculations, we assume a constant transmission model, where the data transmission rate remains uniform during a complete collaborative task assignment. The experimental parameters and their corresponding values are detailed in Table \ref{tab:table2} \cite{wang2024computation}.

\textit{2) Simulation Scenario}

We perform simulations in the following scenario: an airship positioned at an altitude of 10 km, accompanied by nine UAVs at an altitude of 3 km, covering an area of 12×12 km². The airship is located at the center of this area, whereas the leader UAV and follow UAVs are uniformly distributed throughout it. The airship obtains task directives from the ground control center, formulates the optimal flight sequence for the designated task targets, and transmits the planned route information to the lead aircraft. The lead aircraft, along with several wingmen UAV, then executes a coordinated flight along the predetermined route. It is important to note that all UAVs fall within the coverage area of the airship and are interconnected and capable of communicating with one another. Furthermore, considering overall accuracy in detecting target amounts, the Yolov5 model outperforms the seven attention-optimized Yolov5 variants, as well as other state-of-the-art object detection models \cite{xiong2024detecting}. In this study, we employ Yolov5 as the task model and perform experiments to evaluate the effectiveness of the proposed approach.

\captionsetup[subfloat]{font=small} 

\textit{3) Benchmark Algorithms}

To demonstrate the advantages of the proposed method, we examine the following benchmark algorithms.

\textbf{Greedy algorithm} involves making the best choice at each step based on the current state, with the aim of achieving a global optimal solution through a series of local optimal solutions. 

\textbf{MADDPG} use a centralized training and distributed execution framework. It extends single agent RL algorithms from the actor-critic framework to accommodate multi-agent scenarios \cite{qin2023multi}. 

\textbf{MADDPG with path planning} integrates the proposed path planning approach with MADDPG, facilitating the generation of task assignment decisions that align with the requirements for task execution.


\begin{table}[!t]
	\caption{ Experiment parameters\label{tab:table2}}
	\centering
	\begin{tabular}{|c|c|}
		\hline
		Parameters & Values \\
		\hline
		Number of UAV $N$ & $[5,10]$ \\
            \hline
		Number of DNN types $\mathbb{K}$ & $[1,3]$  \\
		\hline
		Number of target coordinates $\mathbb{W}$ & $[10,50]$  \\
		\hline
	    Size of task $\mathbb{L}_\mathbb{W}$& $[0,80]$ GB \\
		  \hline
		Number of random $k$ & $5$  \\
  	  \hline
		Weight of distance $\vartheta$ & 0.5  \\
  	  \hline
		Weight of task size $\rho$ & 0.5  \\
            \hline
		UAV propulsion speed $\nu$ & 20m/s  \\
            \hline
		Average task processing rate $\upsilon_{\mathrm{q}}$ & 10GB/min  \\
            \hline	
		Computational complexity $c$ & $[50,500]$ cycle/bit \\
		\hline
		Maximum transmission power $P_{n}$  &$[50,100]$ mW \\
		\hline
	    Maximum bandwidth ${B_n}$ &$[1,5]$ MHz \\
		\hline	
		Maximum cache resources $m$ & $[100,500]$ GB \\
		\hline	
		Maximum tolerable latency $\tau$ & $[10,200]$ ms \\
		\hline
		Noise power $P_{N}$ & -115 dBm \\
		\hline
		  Computing capacity of UAV $f_{n}$ & 15 Gigacycles/s \\

		\hline
	\end{tabular}
\end{table}


 \begin{figure*}[t!]
 	\centering
 	  \subfigure[]{
 \includegraphics[width=0.31\textwidth]{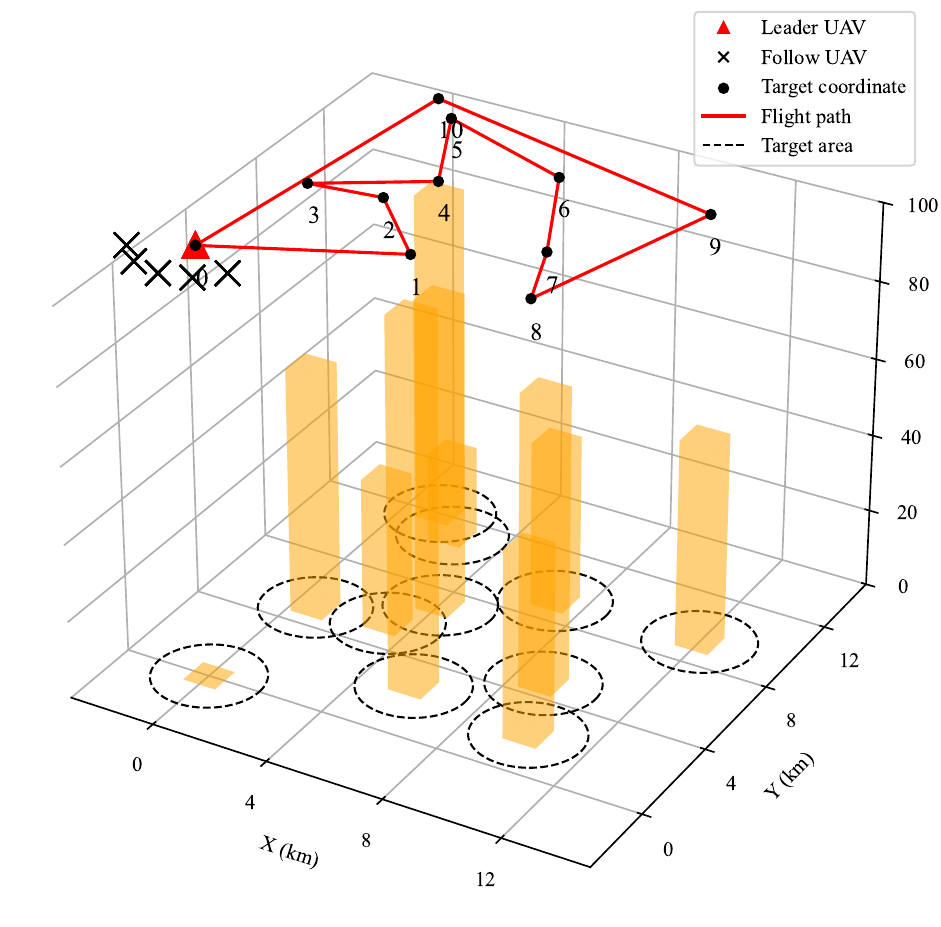}
 	}
        \hfill
         \subfigure[]{
 \includegraphics[width=0.31\textwidth]{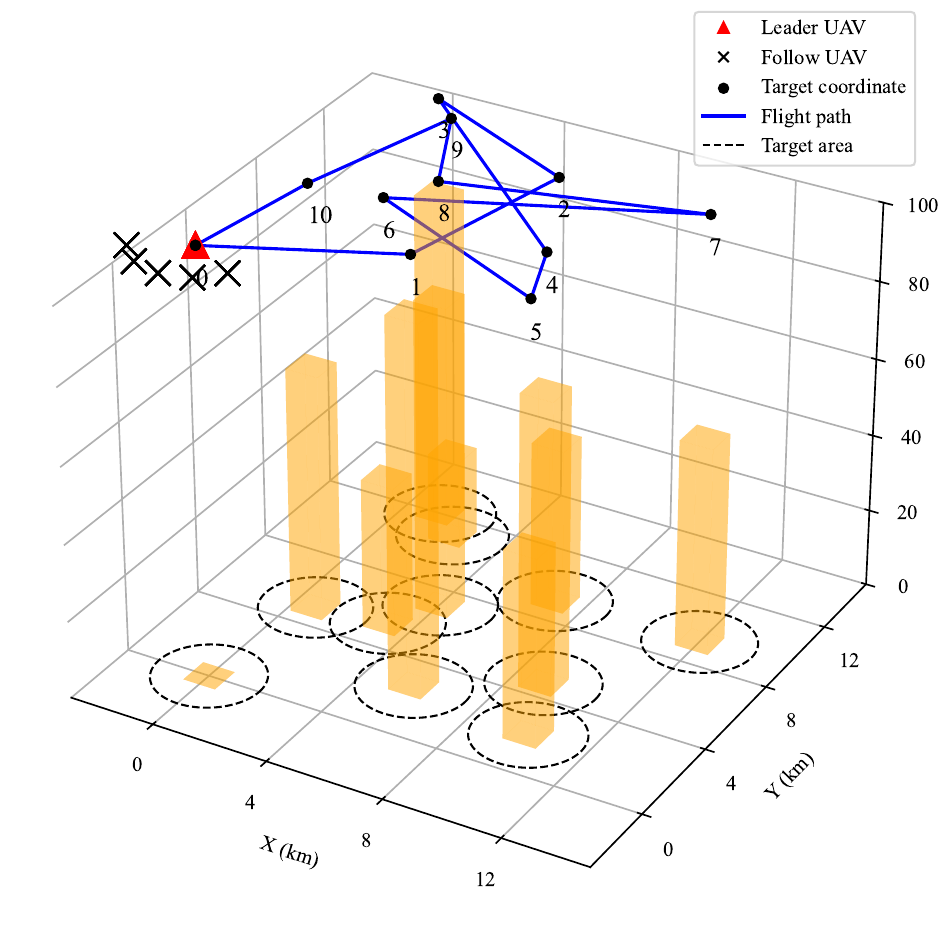}
 	}
        \hfill    
         \subfigure[]{
 \includegraphics[width=0.31\textwidth]{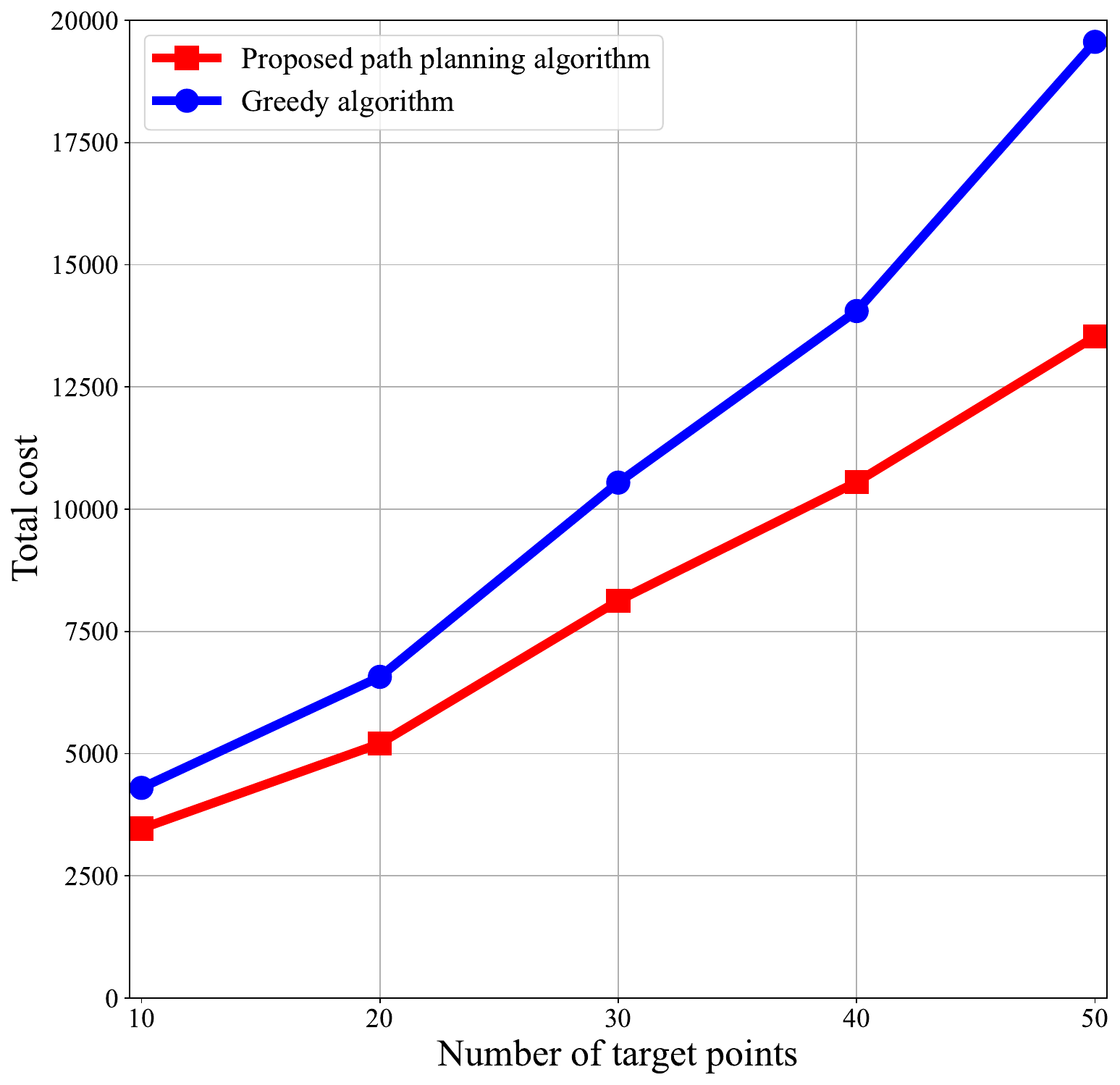}
 	}
 	\caption{UAV path planning considering task size and flight distance.}
 \end{figure*}

\subsection{Performance Evaluation}
\textit{1) Path Planning}

This experiment simulated a UAV inspection area consisting of ten target coordinates, each defined by a distinct task size. Fig. 5(a) shows the proposed UAV path planning method. By comprehensively considering both task size and flight path, this method generates the global optimal path and target coordinate sequence, prioritizing target coordinates with larger task sizes while ensuring coverage of all target coordinates. In contrast, Fig. 5(b) presents a greedy algorithm which selects the best option at each step based solely on the current state, and is prone to local optima. As a result, it can lead to a significant increase in flight path length, reducing the survival time of the UAV swarm and hindering its ability to conduct long-term operations under energy constraints. 


 \begin{figure}[t!]
 	\centering
 	  \subfigure[]{
 \includegraphics[width=0.21\textwidth]{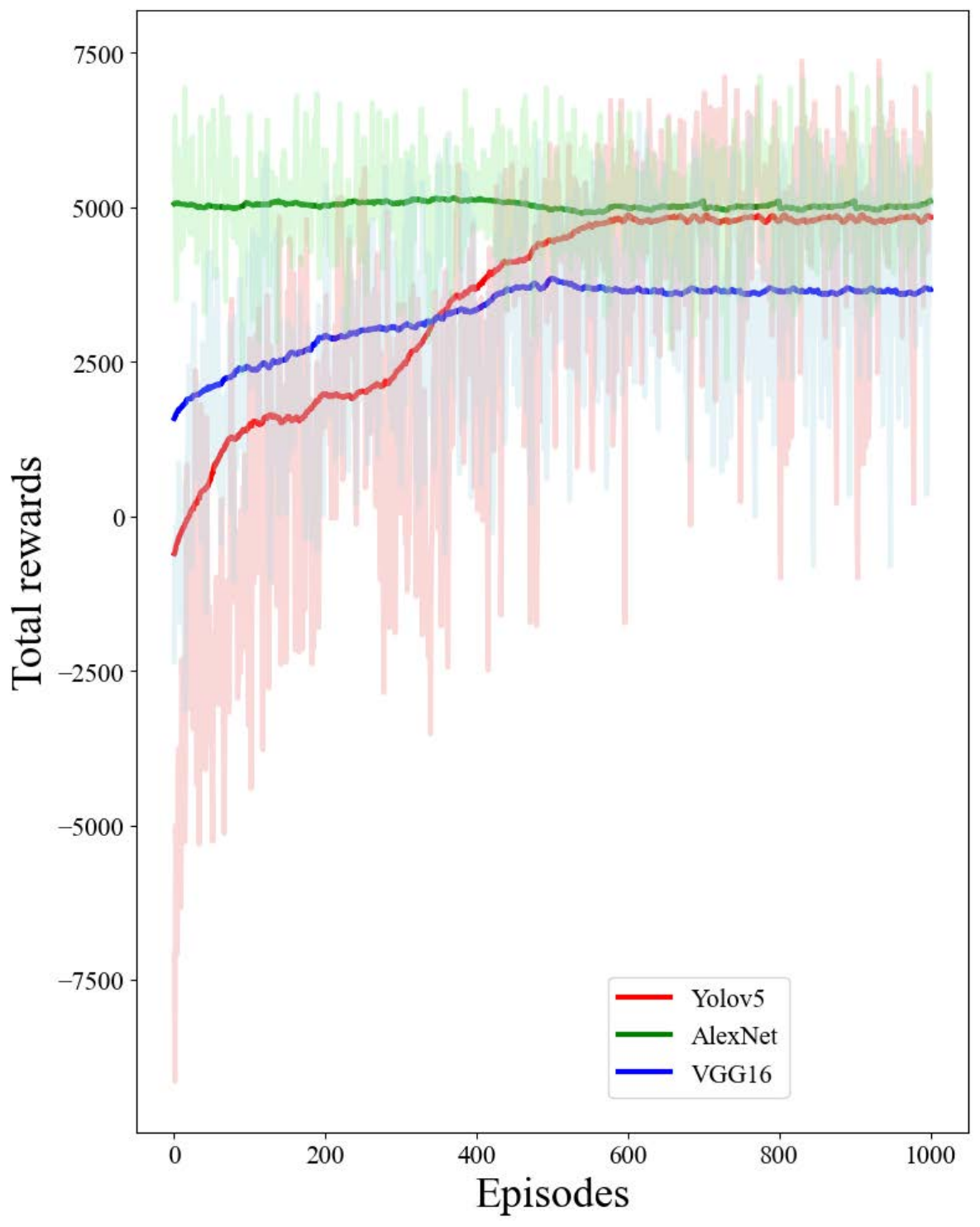}
 	}
        \hspace{0.025\linewidth}  
         \subfigure[]{
 \includegraphics[width=0.21\textwidth]{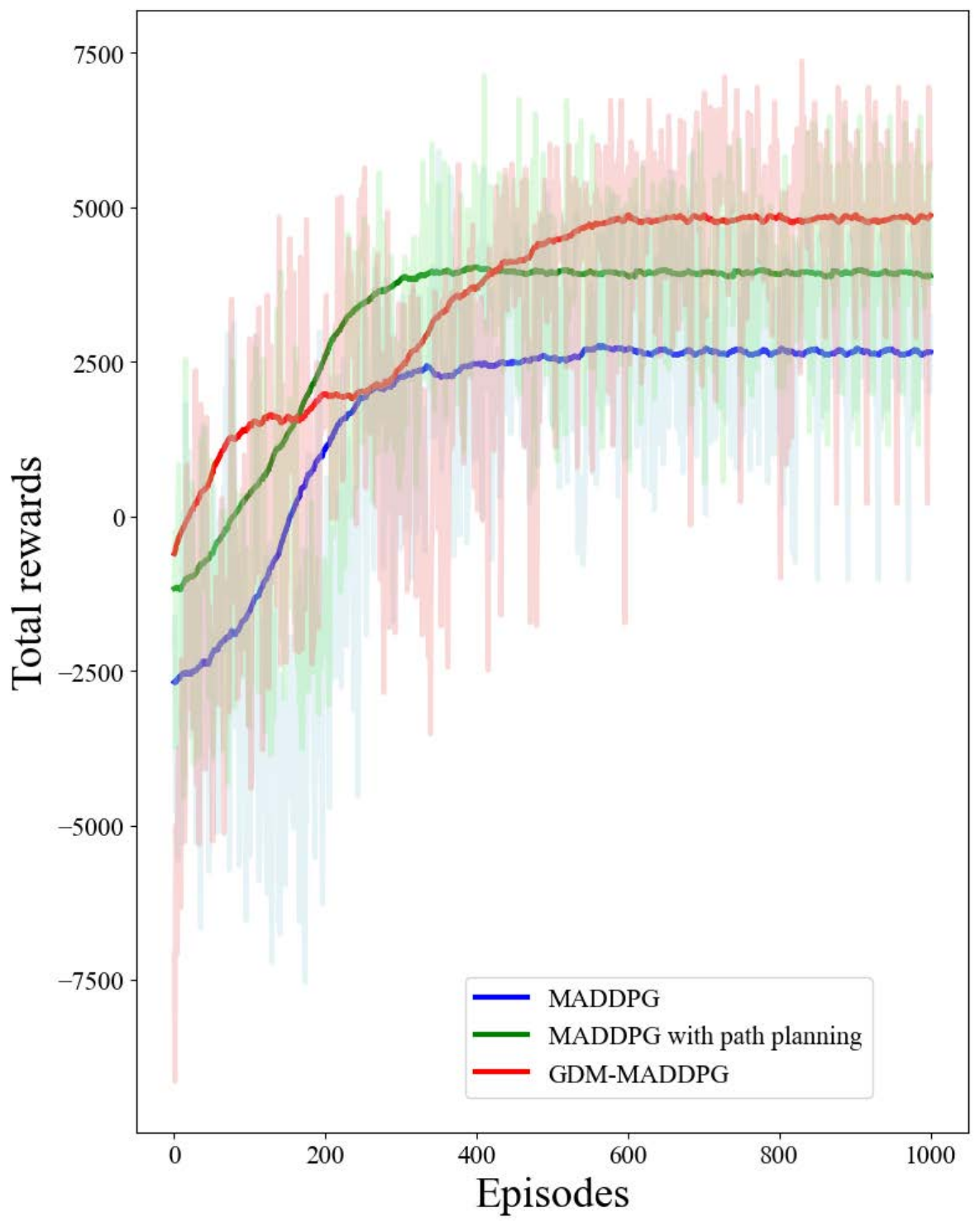}
 	}
 	\caption{The training of different DNN models and benchmark algorithms.}
 \end{figure}

As illustrated in Fig. 5(c), the cost associated with the optimal path finding of the UAV increases as the number of target coordinates increases. In particular, the path planning algorithm proposed in this study consistently exhibits lower costs compared to the greedy algorithm in all scenarios evaluated. Specifically, for target coordinate counts of 10, 20, 30, 40, and 50, the total cost produced by the proposed algorithm is reduced by approximately 19.3\%, 21.1\%, 23.0\%, 25.6\%, and 27.1\%, respectively, compared to the greedy algorithm. This enhancement can be attributed to the expanding solution space that accompanies an increase in the number of target coordinates. As the candidate set of target coordinates grows, the proposed algorithm has greater opportunities to explore solutions that yield lower total costs, ultimately facilitating the identification of the optimal path with minimal total cost. Thus, it is evident that the algorithm developed in this study is particularly adept at addressing path planning challenges that involve a larger number of target coordinates.

\textit{2)Training Procedures}

The convergence of GDM-MADDPG in various tasks of the DNN model is illustrated in Fig. 6(a). Despite the differences in structural complexity between DNN models, the proposed method successfully accommodates these tasks and achieves convergence. The AlexNet model, characterized by fewer layers and lower computational requirements, demonstrates favorable convergence performance. In contrast, the VGG16 model, with its greater number of layers and higher computational demands, experiences increased processing delays, resulting in lower rewards and reduced convergence efficacy for our method. Although the convergence performance of our method on the YOLOv5 model is slightly inferior to that of AlexNet, YOLOv5, as a widely adopted real-time object detection algorithm, has a model complexity that lies between the two aforementioned models. Consequently, subsequent experiments employ YOLOv5 as the experimental task model to validate the superior performance of this method in terms of AOI and task completion rate.

We perform ten independent training runs for each algorithm. As illustrated in Fig. 6(b), although the proposed GDM-MADDPG converges at a slower rate than the two algorithms, GDM-MADDPG demonstrates superior performance compared to the other solutions. The total reward attained by GDM-MADDPG increases throughout the training process, eventually converging to approximately 5000 after around 600 training episodes. This enhancement is attributed to the generative capabilities of GDM, which markedly improve action sample efficiency by progressively reducing noise through multiple denoising steps. Finally, the low performance of the two algorithms further highlights the effectiveness of the proposed GDM-MADDPG method.


 \begin{figure}[t!]
 	\centering
 	  \subfigure[]{
 \includegraphics[width=0.21\textwidth]{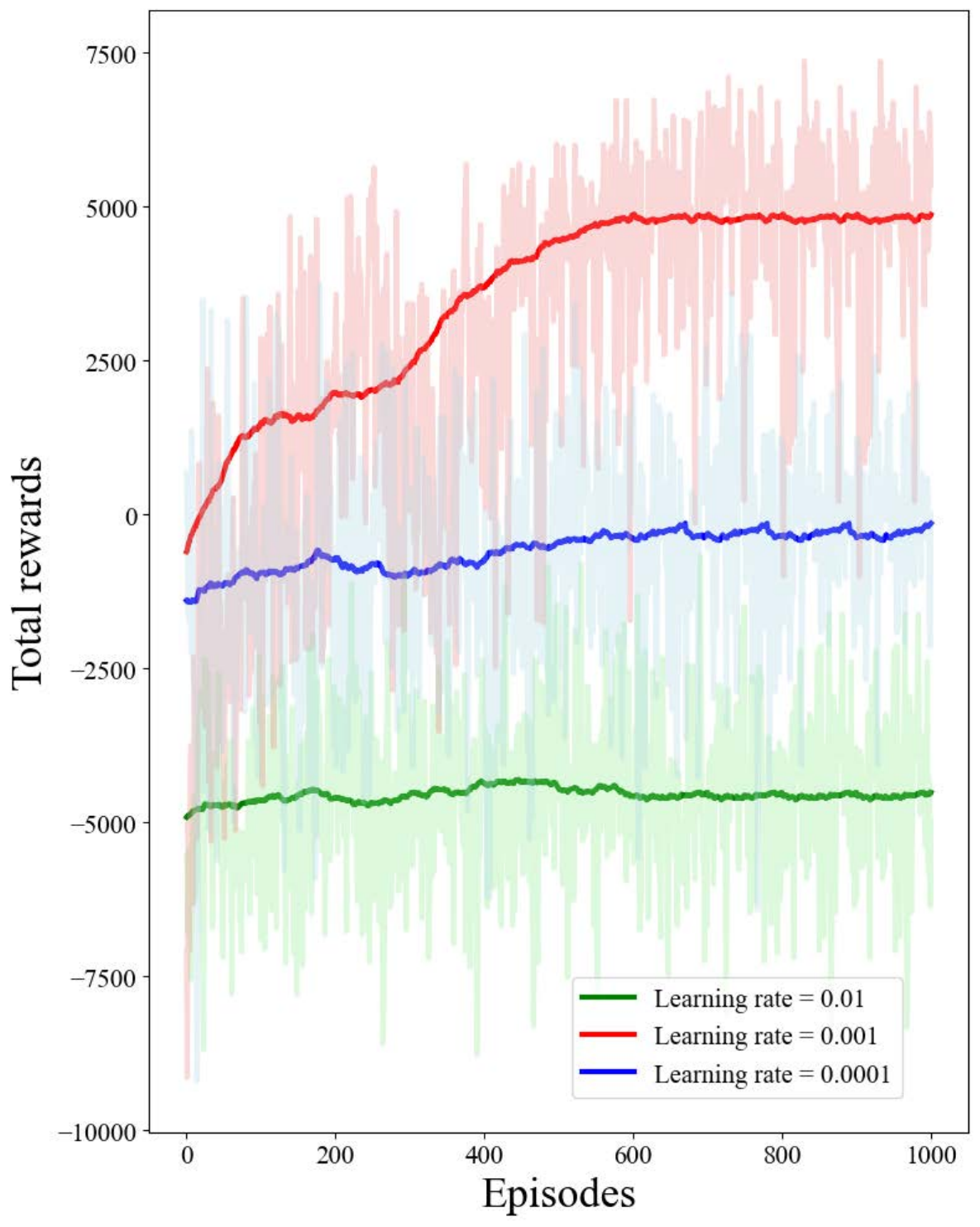}
 	}
        \hspace{0.025\linewidth}  
         \subfigure[]{
 \includegraphics[width=0.21\textwidth]{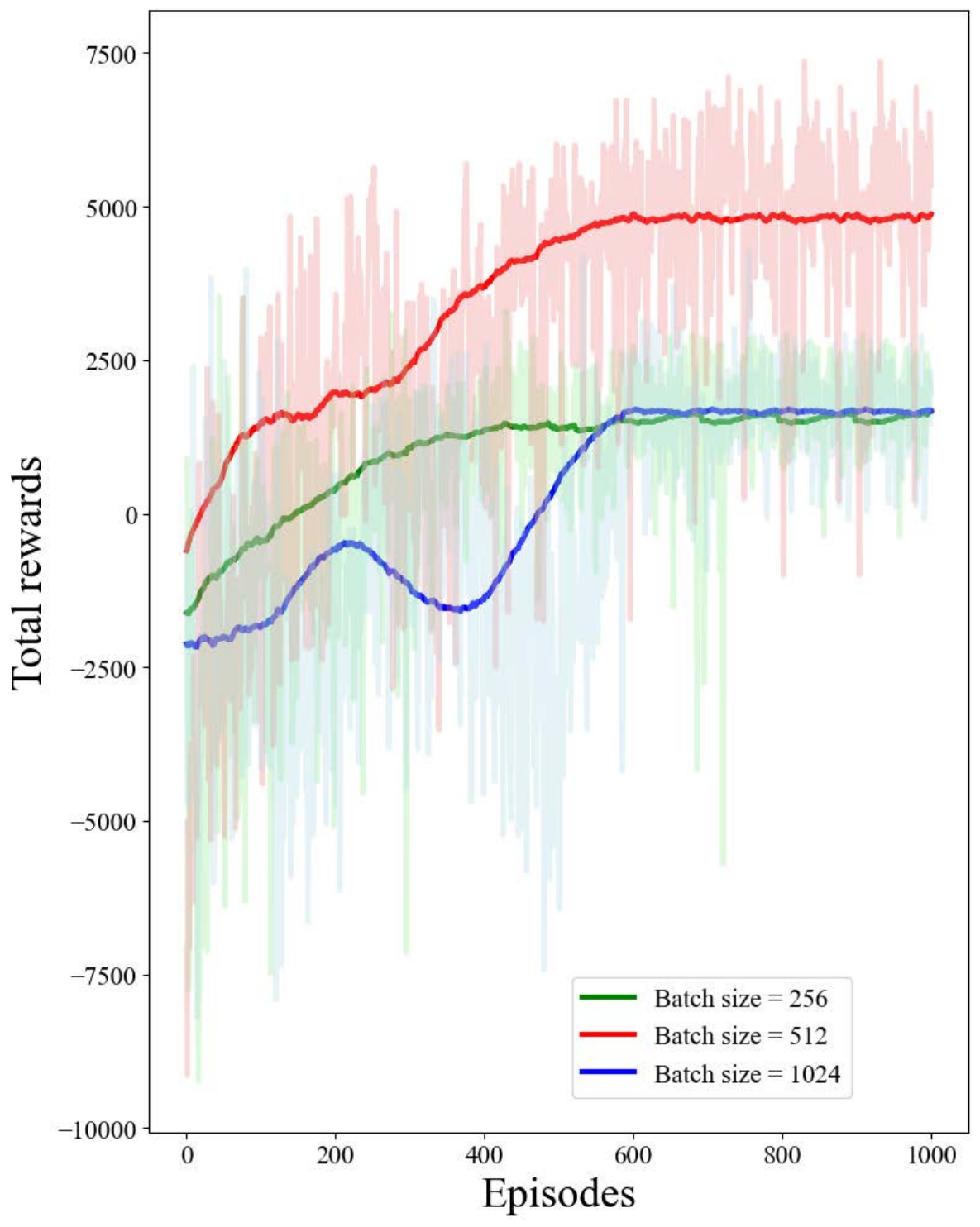}
 	}
 	\caption{The training of different learning rate and batch size.}
 \end{figure}

By conducting a series of systematic experiments, we aim to identify the optimal values for two key parameters that influence the performance of the MADDPG algorithm: 1) learning rate and 2) batch size. The experimental results, illustrated in Fig. 7(a), clarify the convergence behavior of the reward function under varying learning rates. A learning rate of 0.01 is excessively large, preventing convergence even after 1000 training iterations. Similarly, learning rates of 0.0001 do not facilitate convergence. In particular, the 0.001 learning rate promotes faster convergence. The suitable learning rate of 0.001 is used to assess the effects of different batch sizes on training performance. As illustrated in Fig. 7(b), a batch size of 512 is recognized as the most advantageous option.

\textit{3) Performance Comparison}

This study compares the AoI of different methods in varying task sizes, as illustrated in Fig. 8(a). The AoI performance of the MADDPG-based method is significantly improved when combined with the path-planning approach. Furthermore, the GDM-MADDPG method proposed in this paper, which is based on MADDPG and incorporates the advantages of GDM in handling complex data distributions and generating high-quality task assignment decisions, improves the timeliness and generalizability of task assignment decision generation. This results in the generation of the most suitable task assignment decisions for the current state, thereby reducing the transmission time during serial task execution. According to Eq. \eqref{E002}, this method also addresses the optimization of path planning for target coordinates and tasks, which reduces waiting times associated with tasks and improves AoI.

The task completion rates of different methods were compared in various task sizes, as shown in Fig. 8(b). The method proposed in this paper outperforms other methods in optimizing system utility, indicating that optimized task path planning and task offloading decision generation achieve a joint optimization of multiple objectives. In terms of path planning, the method not only considers the optimal path but also accounts for the task-size requests at the target coordinates, enhancing the selectivity of local optima. Regarding DNN task assignment, factors such as task partitioning decisions and load balancing are incorporated. Additionally, the introduced GDM reverse denoising process improves the learning and generation capabilities of agent decisions, leading to well-generalized action decisions. Consequently, as the task size gradually increases, the MADDPG-based decision method lacks robust dynamic adaptability under multiple constraints. Its learning capability is constrained by the fixed learning mechanism of the actor-critic network, resulting in the absence of a strategy with generalization capabilities. 


 \begin{figure*}[t!]
 	\centering
 	  \subfigure[]{
 \includegraphics[width=0.31\textwidth]{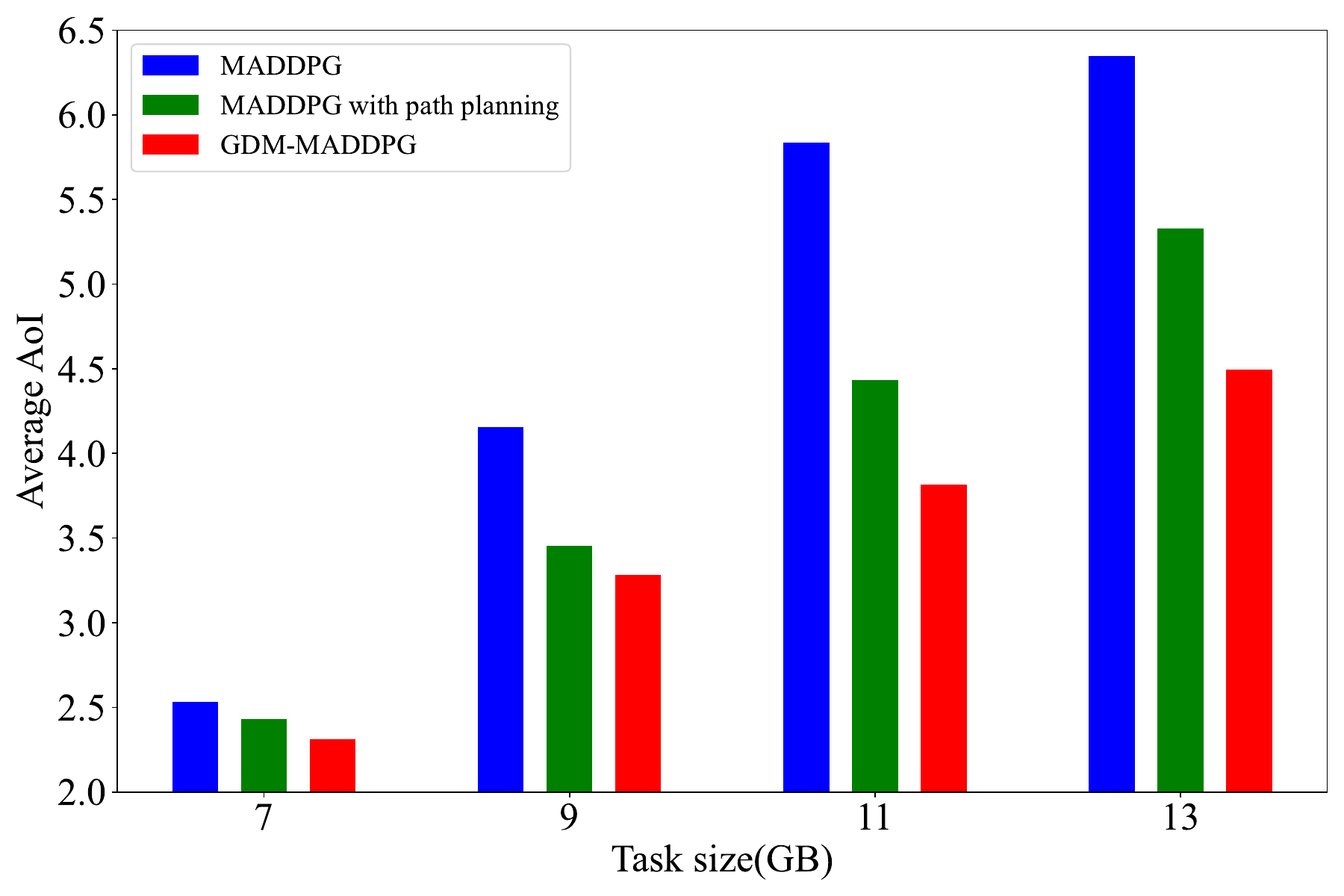}
 	}
        \hfill
         \subfigure[]{
 \includegraphics[width=0.31\textwidth]{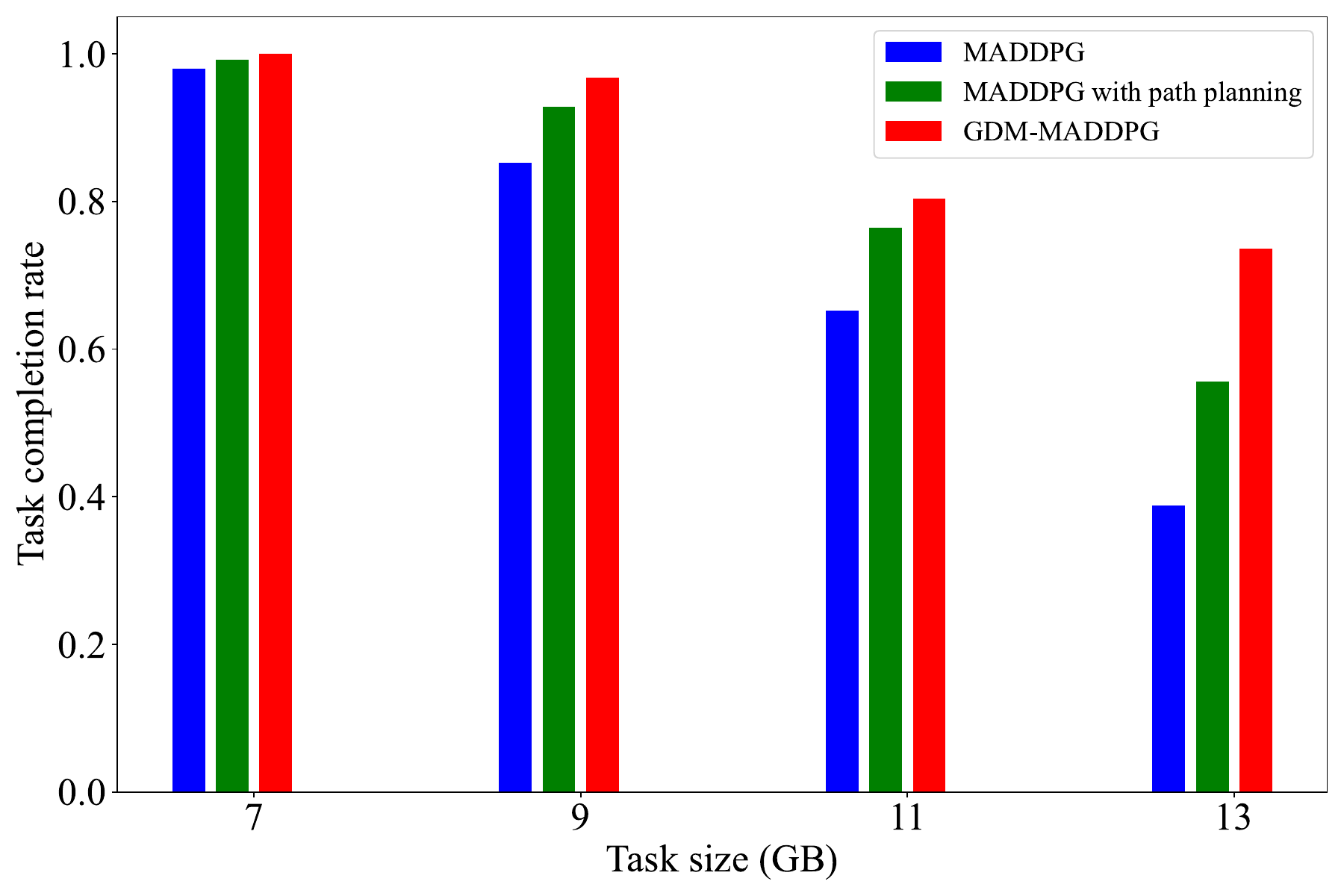}
 	}
        \hfill    
         \subfigure[]{
 \includegraphics[width=0.31\textwidth]{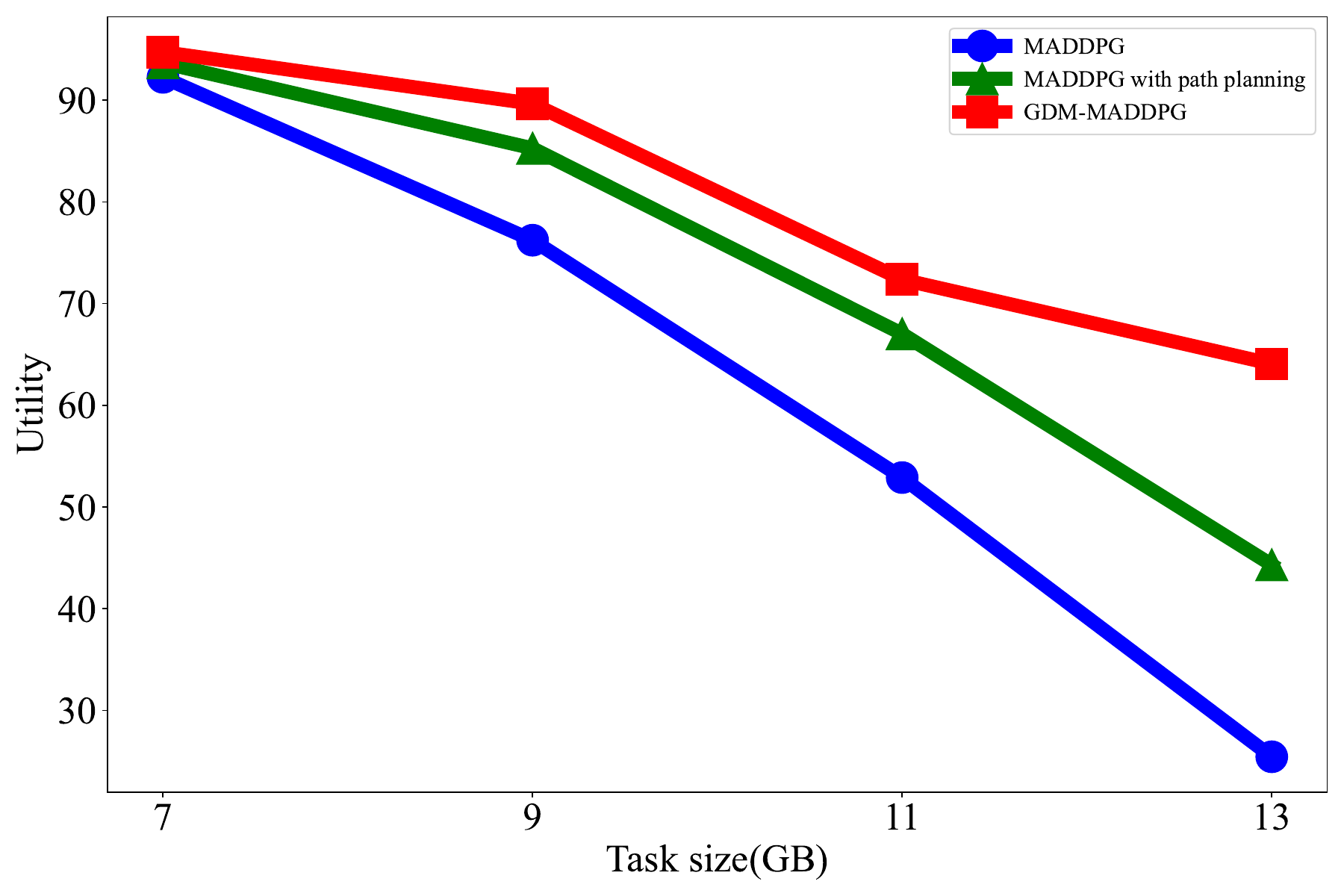}
 	}
 	\caption{Comparison of the AoI, task completion rate, and utility across varying task sizes.}
 \end{figure*}

The utility of different methods was compared under various task sizes, as shown in Fig. 8(c), as task sizes increase, the need for additional task assignment work is incurred under latency constraints, directly increasing the latency of task completion. This results in a downward trend in task completion rates for all methods. The proposed method considers task size constraints during the path planning stage, and task assignment decisions can be rapidly learned and generated through the GDM method, further reducing task processing delays and allowing a greater number of tasks to be completed within the specified time. In addition, these findings indicate that the proposed method is suited for large-scale tasks.

\section{Conclusion}
\label{sec:8}
In this paper, we addressed the challenge of planning joint flight paths, assigning DNN tasks, and balancing load in UAV networks as a two-stage optimization problem. In the first stage, we consider the task size of the area to be inspected and the shortest flight path as optimization constraints. We then employ a novel heuristic algorithm to optimize the UAV's flight path, focusing on maximizing system utility. In the second stage, we introduce a novel GDM-MADDPG algorithm to determine optimal DNN task assignment decisions. Through numerical simulations, we demonstrate the superior performance of our proposed algorithm compared to existing benchmark solutions.

Research in the future could explore the feasibility of employing an expert dataset, generated offline through a brute-force search method, to enhance the forward process of GDM. Subsequently, supervised learning could be employed to train GDM to align with the action distribution produced by the reverse process and the expert data. Furthermore, exploring DNN task assignment for emergency response and post-disaster search scenarios is also a valuable research area.


\bibliographystyle{IEEEtran}

\end{document}